\pdfoutput=1

\documentclass[preprint,onecolumn]{revtex4-2}

\usepackage{graphicx}
\usepackage{siunitx}
\usepackage[hidelinks]{hyperref}
\usepackage{gensymb}
\usepackage{float}
\usepackage{lineno}
\usepackage{xcolor}
\usepackage{amsmath, amsthm, amssymb}
\usepackage{mathrsfs}

\usepackage{microtype}
\usepackage{caption}
\usepackage{subcaption}

\usepackage[ruled,vlined,linesnumbered]{algorithm2e}

\usepackage[british]{babel}

\usepackage{color}
\usepackage{tabularx} % Required for tabularx
\usepackage{booktabs} % For better-looking tables

\definecolor{mygreen}{HTML}{009B55}

\graphicspath{ {figures/} }

\begin{document}

\title{Compositional neural networks for multimodal X-ray and neutron scattering inversion}

\author{Valentin Munteanu\textsuperscript{a}}
\email{valentin.munteanu@uni-tuebingen.de}

\author{Alexander Hinderhofer\textsuperscript{a}}
\author{Dmitry Lapkin\textsuperscript{a}}
\author{Vladimir Starostin\textsuperscript{b}}

\author{Frank Schreiber\textsuperscript{a}}
\email{frank.schreiber@uni-tuebingen.de}

\affiliation{
[a] Institute of Applied Physics, University of Tübingen, Auf der Morgenstelle 10, 72076 Tübingen, Germany
}
\affiliation{
[b] Cluster of Excellence "Machine Learning" - new perspectives for science,  University of Tübingen, Germany
}

\begin{abstract}
X-ray and neutron scattering are indispensable for structural characterization across condensed-matter physics and materials science. Yet, because scattering experiments generally record intensities without phase information, distinct structures can yield similar scattering patterns, rendering the corresponding inverse problems non-unique. This ambiguity can be reduced by combining complementary measurements, but their integration into machine-learning frameworks remains challenging. Here we introduce a modular framework for probabilistic multi-measurement analysis based on prior-amortized flow matching. Single-observation posterior estimators are trained separately for individual observation types or physical probes and combined only at inference time through a compositional sampler. This late fusion approach allows pretrained single-observation models to be reused without dedicated joint retraining. As a case in point, we consider reflectometry, which provides depth-resolved, nanometer-scale information about thin films and interfaces. Our single-observation demonstrations span X-ray, time-of-flight neutron and polarized neutron reflectometry. In separate co-refinement studies, task-matched estimators enable homogeneous composition across multiple neutron contrasts and heterogeneous multimodal composition of X-ray and neutron measurements. Together, these results establish compositional flow matching as a modular approach to uncertainty-aware inversion and co-refinement, with potential applicability to other scattering problems in which multiple observations constrain a common physical structure while retaining measurement-specific parameters.
\end{abstract}

\maketitle
% \linenumbers
\renewcommand\thefigure{\arabic{figure}}
\renewcommand\thetable{\arabic{table}}

\clearpage

\section*{Introduction} 
X-ray and neutron scattering are central to structural characterization in condensed-matter physics and materials science~\cite{AlsNielsen2011,Tolan1999X}. Since scattering experiments generally record intensities without phase information, recovering the underlying structure can lead to non-unique or only weakly constrained inverse problems. Distinct structures or parameter combinations can therefore yield similar scattering patterns, so a single best-fitting structure may conceal alternative solutions that remain compatible with the data.
Two distinct strategies address this ambiguity. First, probabilistic inversion uses Bayesian inference to characterize a posterior distribution over structural and instrumental parameters rather than returning a single structure. Second, measurements with complementary sensitivities can be combined so that parameters weakly identified by one observation may be constrained by another.

Posteriors arising from scattering inverse problems can be strongly correlated, non-Gaussian or multimodal, and joint inference enlarges the parameter space by introducing both quantities shared across measurements and quantities specific to an individual contrast, probe or instrument. These characteristics motivate flexible posterior approximations that can represent complex uncertainty across a range of possible observations. Simulation-based inference (SBI) ~\cite{Cranmer2020} provides such an approach by training neural networks on simulated parameter--observation pairs to approximate posterior distributions. In the amortized setting, the main computational burden is incurred during training, after which inference for new measurements can be performed rapidly. This makes SBI attractive for repeated or high-throughput analysis in experimental science, including at X-ray and neutron facilities.

Amortized posterior estimators are nevertheless typically trained for a fixed observation model, and a network trained for one simulator cannot in general be combined directly with a network trained for another. Training a new joint model for every number and combination of measurements would undermine the modularity and computational reuse offered by amortization. Compositional SBI provides a foundation for combining posterior information from multiple observations at inference time~\cite{geffner2023compositional,linhart2026diffusionposteriorsamplingsimulationbased}. Many joint inversion problems, however, require an additional structure of shared and local parameters. Some parameters describe a common physical system and must be shared across observations, whereas others describe a particular contrast, probe or instrument and remain local to an individual measurement. A modular multi-measurement framework must therefore combine separately trained estimators in a common parameter space while correcting the repeated contribution of the shared prior. 

Flow matching~\cite{lipman2023flow,tong2024improving} provides a convenient basis for such a compositional construction. It trains a time-dependent vector field that transports samples from a simple base distribution to the target distribution along a prescribed probability path. Flow matching posterior estimation (FMPE) has been shown to be a competitive and scalable SBI method~\cite{Wildberger_NeurIPS_FlowMatchingSBI}. For the probability path used here, the learned vector field also determines a posterior score, which enables the information supplied by separately trained posterior estimators to be composed during sampling.

Specular X-ray and neutron reflectometry are indispensable techniques for resolving the depth-dependent structure and properties of thin films, multilayers and buried interfaces~\cite{Daillant2009X,Ankner1993Neutron}. Beyond its broad experimental relevance, reflectometry presents a demanding setting for this modular inference framework because it spans different probes, acquisition modes and observation structures. X-ray reflectometry (XRR) is sensitive to the electronic scattering length density (SLD), neutron reflectometry (NR) is sensitive to the nuclear SLD and can exploit isotope-dependent contrast, and polarized neutron reflectometry (PNR) further extends this sensitivity to magnetic properties through the spin dependence of the neutron interaction~\cite{Majkrzak1991Polarized,Causer2023}. Joint analysis of such measurements can constrain parameters that are weakly identified by an individual observation. The information gained nevertheless depends on the experimental design, measurement strategy, contrast engineering and model parameterization~\cite{Treece2019Optimization,McCluskey2020general,Durant2021Determining,Holt2009}. In practical reflectometry analysis, prior knowledge about plausible thickness ranges, expected material SLDs or instrumental conditions further constrains the solution space. 

XRR is performed at synchrotron light sources~\cite{Schroer2022,Haensel1992} or laboratory sources, whereas NR uses monochromatic or time-of-flight (TOF) reflectometers~\cite{Campbell2011}. Energy-dispersive XRR and rotating-sample XRR geometries further illustrate the move towards rapid and parallel acquisition~\cite{Kowarik2007EDXRR,Lippmann2016FastXRR}. Such developments increase the relevance of amortized analysis for in situ and time-resolved scattering experiments~\cite{Ritley2001,Kowarik2006Real}. PNR additionally requires polarization components such as polarizers, spin flippers and, depending on the experimental configuration, spin analyzers. In full spin-resolved PNR, four reflectivity cross sections are distinguished: the non--spin-flip channels \(R^{\uparrow\uparrow}\) and \(R^{\downarrow\downarrow}\), and the spin-flip channels \(R^{\uparrow\downarrow}\) and \(R^{\downarrow\uparrow}\), where the first and second arrows denote the incident and reflected neutron-spin states, respectively. Magnetization components parallel to the neutron-spin quantization axis contribute to the non--spin-flip channels, whereas components transverse to this axis couple the two spin states and generate spin-flip scattering. When the incident neutron polarization is selected but the reflected spin state is not resolved, the two measured intensities sum over the possible reflected spin states, such that \(R^{\uparrow}=R^{\uparrow\uparrow}+R^{\uparrow\downarrow}\) and \(R^{\downarrow}=R^{\downarrow\downarrow}+R^{\downarrow\uparrow}\). For magnetization collinear with the neutron-spin quantization axis, the spin-flip cross sections vanish, and the measured intensities reduce to \(R^{\uparrow}=R^{\uparrow\uparrow}\) and \(R^{\downarrow}=R^{\downarrow\downarrow}\), as assumed by the two-channel PNR model used for the experimental co-refinement examples in this work. Because these channels have fixed physical labels and ordering, they are treated as a single multichannel observation rather than as separate observations combined through compositional inference. 

Previous machine-learning approaches to reflectometry have included point-estimator neural networks for XRR and NR and models incorporating prior knowledge~\cite{Greco2019Fast,Greco2021Neural,Greco2022Neural,Aoki2021Deep,Mironov2021,CarmonaLoaiza2021,Doucet2021,Munteanu2024}. Other studies have addressed high-throughput and autonomous-experiment settings~\cite{Pithan2023, Marecek2022,SchumiMareckeck2025}, as well as variational-autoencoder analyses of PNR data~\cite{Andrejevic2022Elucidating}. Probabilistic approaches based on mixture-density networks or normalizing flows have further enabled uncertainty-aware single-observation inference~\cite{Kim2021,vladimir_panpe}. In particular, prior-amortized neural posterior estimation (PANPE) conditions the generative model not only on the measured data but also on user-provided prior bounds, allowing experiment-specific prior information to be changed at inference time without retraining~\cite{vladimir_panpe}. These developments make fast single-observation inference increasingly realistic. However, they do not provide a modular mechanism for combining estimators trained separately for different measurements, experimental contrasts or physical probes. 

Here we introduce a prior-amortized compositional flow matching framework for probabilistic reflectometry analysis. This work makes two closely connected contributions. First, we adapt prior-amortized flow matching posterior estimation to a broad range of experimentally relevant XRR, NR and PNR settings. Second, building on recent compositional SBI approaches~\cite{linhart2026diffusionposteriorsamplingsimulationbased}, we introduce compositional posterior sampling for reflectometry and extend the formalism to accommodate shared and local parameter blocks. Specifically, we derive the composition equations for separately trained estimators whose parameter spaces comprise a common block of shared parameters together with observation-specific local parameters. The resulting sampler combines compatible posterior estimators trained separately for individual observations or physical probes. The user specifies which parameters are shared and which remain local, the score contribution from each estimator is embedded into the resulting joint parameter space, and the repeated contribution of the shared prior is corrected during composition. In reflectometry, shared parameters can include layer thicknesses and interfacial roughnesses, whereas contrast-dependent SLDs, backgrounds, scale factors and probe-specific scattering properties may remain local to individual observations. This supports homogeneous composition, in which the same trained model is reused across several observations with a common parameterization as in multi-contrast NR, and heterogeneous composition, in which separately trained XRR and (P)NR models are coupled through shared structural parameters. Throughout this work, \emph{multimodal} refers to the combination of different experimental modalities or probes. Although demonstrated here for reflectometry, the shared--local compositional construction is applicable more broadly to X-ray and neutron scattering problems in which multiple observations constrain a common physical model while retaining measurement-specific parameters.

The remainder of the paper is organized as follows. We first demonstrate the approach on a sequence of reflectometry problems of increasing complexity, progressing from single-curve XRR and NR inference to multichannel PNR, multi-contrast NR co-refinement and multimodal XRR/PNR co-refinement. We then discuss the scope and limitations of the framework. The mathematical formulation, neural-network architecture, compositional sampler and training-data generation are provided in the Methods.

\clearpage

\section*{Results}
\label{sec:applications}

\subsection*{Bayesian compositional inference with shared and local parameters}

We developed a modular Bayesian analysis framework in which posterior estimators trained for individual observation models can be reused and composed at inference time. For each observation model, a prior-amortized flow matching posterior estimator is trained on simulator-generated pairs \(\{(\theta_i,C_i)\}_{i=1}^{N}\). Here, \(C=(y,b,\eta)\) denotes the complete conditioning context, comprising
the measured data \(y\), user-supplied bounds \(b\) defining the prior domain
and, where relevant, known acquisition quantities \(\eta\). The trained estimator approximates the posterior \(p(\theta\mid C)\) for admissible contexts within the support of its training distribution.

Flow matching approximates the conditional posterior by learning a time-dependent vector field that transports a Gaussian base distribution toward the target posterior distribution. Posterior samples are generated by drawing an initial state from the base distribution and integrating \begin{equation} \frac{\mathrm d\theta_t}{\mathrm dt} = v(\theta_t,t;C) \label{eq:ode_main} \end{equation} from \(t=0\) to \(t=1\). For the Gaussian probability path used here, the learned vector field can be converted at each intermediate time into the score \( s_{t}(\theta_t;C) = \nabla_{\theta_t}\log p_t(\theta_t\mid C) \) of the corresponding path distribution. Access to this score representation enables separately trained posterior estimators to be composed during inference.

We use composition when several measurements probe the same physical sample. Consider, for example, the multimodal co-refinement of an XRR and an NR measurement from the same structure. Parameters collected in \(\theta^{\mathrm{sh}}\), such as layer thicknesses and interfacial roughnesses, describe structure shared by both probes. The remaining parameters, \(\theta^{\mathrm{loc}}_1\) and \(\theta^{\mathrm{loc}}_2\), are specific to one measurement, such as probe-dependent scattering-length densities or instrumental quantities. Assuming conditionally independent measurement likelihoods and compatible factorized priors with a common shared-parameter marginal, the joint Bayesian posterior can be rewritten in terms of the two single-observation posteriors represented by the trained estimators as
\begin{equation}
p\!\left(
\theta^{\mathrm{sh}},
\theta^{\mathrm{loc}}_1,
\theta^{\mathrm{loc}}_2
\mid C_1,C_2
\right)
\propto
\frac{
p_1\!\left(\theta^{\mathrm{sh}},\theta^{\mathrm{loc}}_1\mid C_1\right)
p_2\!\left(\theta^{\mathrm{sh}},\theta^{\mathrm{loc}}_2\mid C_2\right)
}{
p\!\left(\theta^{\mathrm{sh}}\mid b^{\mathrm{sh}}\right)
}.
\label{eq:joint_posterior_two_main}
\end{equation}
The two posteriors jointly constrain the shared structure, but each already contains one copy of the shared prior. Dividing by that prior removes the surplus copy so that prior information is counted exactly once. A local parameter block requires no analogous correction because it occurs in only one single-observation posterior.

In the zero-noise limit, taking the logarithmic gradient of Eq.~\eqref{eq:joint_posterior_two_main} gives the score-addition rule
\begin{equation}
\nabla_{\Theta}\log p(\Theta\mid C_1,C_2)
=
\nabla_{\Theta}\log p_1
+
\nabla_{\Theta}\log p_2
-
\nabla_{\Theta}\log p\!\left(\theta^{\mathrm{sh}}\mid b^{\mathrm{sh}}\right),
\label{eq:naive_score_addition_main}
\end{equation}
where each term is embedded in the global coordinates
\(\Theta=(\theta^{\mathrm{sh}},\theta^{\mathrm{loc}}_1,
\theta^{\mathrm{loc}}_2)\). This direct addition is not valid at intermediate path times because Gaussian perturbation does not commute with multiplication of probability densities. The intermediate-time composed score can therefore be written schematically as
\begin{equation}
s_{\mathrm{comp}}(\Theta_t,t;C_{1:2})
=
\nabla_{\Theta_t}\log p_{1,t}
+
\nabla_{\Theta_t}\log p_{2,t}
-
\nabla_{\Theta_t}\log p_t\!\left(\theta_t^{\mathrm{sh}}\mid b^{\mathrm{sh}}\right)
+
s_{\mathrm{corr}}(\Theta_t,t;C_{1:2}),
\label{eq:corrected_score_main}
\end{equation}
where \(s_{\mathrm{corr}}\) accounts for the mismatch between perturbing the joint posterior and combining the individually perturbed posteriors, and vanishes in the zero-noise limit.

We build on the full-covariance \textsc{Gauss} composition method introduced in Ref.~\cite{linhart2026diffusionposteriorsamplingsimulationbased}, which uses Gaussian approximations to account for intermediate-time perturbation during posterior composition. Here, we extend this construction to posterior estimators defined on a common block of shared parameters together with observation-specific local parameter blocks. Retaining the full covariance allows the composition to account both for the information that each observation contributes to the shared parameters and for correlations between shared and observation-specific parameters. The resulting rule performs two coupled operations: it pools the shared score contributions according to their marginal precisions and adjusts each local score according to its correlation with the shared parameters. Let the denoising precision, that is, the inverse covariance of the Gaussian denoising approximation for observation \(j\), be partitioned into shared and local blocks,
\begin{equation}
\Lambda_{j,t}
=
\begin{pmatrix}
\Lambda^{\mathrm{ss}}_{j,t} & \Lambda^{\mathrm{sl}}_{j,t}\\
\Lambda^{\mathrm{ls}}_{j,t} & \Lambda^{\mathrm{ll}}_{j,t}
\end{pmatrix}.
\end{equation}
For two observations, the resulting composed score is
\begin{equation}
\begin{aligned}
s^{\mathrm{sh}}_{\mathrm{comp},t}
&=
\left(
\bar\Lambda_{1,t}+\bar\Lambda_{2,t}-\Lambda^{\mathrm{pr}}_t
\right)^{-1}
\left(
\bar\Lambda_{1,t}s^{\mathrm{sh}}_{1,t}
+
\bar\Lambda_{2,t}s^{\mathrm{sh}}_{2,t}
-
\Lambda^{\mathrm{pr}}_t s^{\mathrm{pr}}_t
\right),
\\
s^{\mathrm{loc}}_{\mathrm{comp},j,t}
&=
s^{\mathrm{loc}}_{j,t}
+
\left(\Lambda^{\mathrm{ll}}_{j,t}\right)^{-1}
\Lambda^{\mathrm{ls}}_{j,t}
\left(
s^{\mathrm{sh}}_{j,t}-s^{\mathrm{sh}}_{\mathrm{comp},t}
\right),
\qquad j=1,2,
\end{aligned}
\label{eq:composed_rule_main}
\end{equation}
where \(s^{\mathrm{pr}}_t\) and \(\Lambda^{\mathrm{pr}}_t\) are the score and denoising precision of the perturbed shared prior, and
\begin{equation}
\bar\Lambda_{j,t}
=
\Lambda^{\mathrm{ss}}_{j,t}
-
\Lambda^{\mathrm{sl}}_{j,t}
\left(\Lambda^{\mathrm{ll}}_{j,t}\right)^{-1}
\Lambda^{\mathrm{ls}}_{j,t}
\label{eq:schur_precision_main}
\end{equation}
is the Schur complement of the local block and represents the marginal precision contributed by observation \(j\) to the shared parameters.

The structure of Eqs.~\eqref{eq:composed_rule_main} and \eqref{eq:schur_precision_main} has a direct physical interpretation. Each single-observation posterior contributes to the shared parameters according to its marginal precision, that is, according to the information carried by the corresponding measurement after uncertainty in its local parameters has been accounted for. The negative prior contribution ensures that prior knowledge enters only once. Conversely, when the joint analysis shifts the shared structure away from the value preferred by one measurement alone, that measurement's local score is adjusted along the shared--local correlation directions of its posterior. Retaining the full
covariance is therefore important whenever structural and probe-specific parameters are correlated. The rule is exact when the individual posteriors and the shared prior are Gaussian and reduces to Eq.~\eqref{eq:naive_score_addition_main} as the perturbation vanishes. The full derivation and its extension to more than two observations are given in the Methods.

The composed score is converted back into a vector field and integrated as in Eq.~\eqref{eq:ode_main} to generate joint posterior samples. We use this construction in two forms. In homogeneous composition, one trained estimator is reused for several observations with a common parameterization, as in multi-contrast NR co-refinement. In heterogeneous composition, separately trained estimators are coupled through a common shared block, as in joint XRR/(P)NR co-refinement. In both cases, the composed sampler returns samples over the shared parameters and every observation-specific local parameter. The mapping of compatible model coordinates into shared and local blocks is specified at inference time, so a dedicated joint estimator does not need to be trained for each supported combination of measurements.

\begin{figure}[h!]
  \centering
  \includegraphics[width=1.0\linewidth]{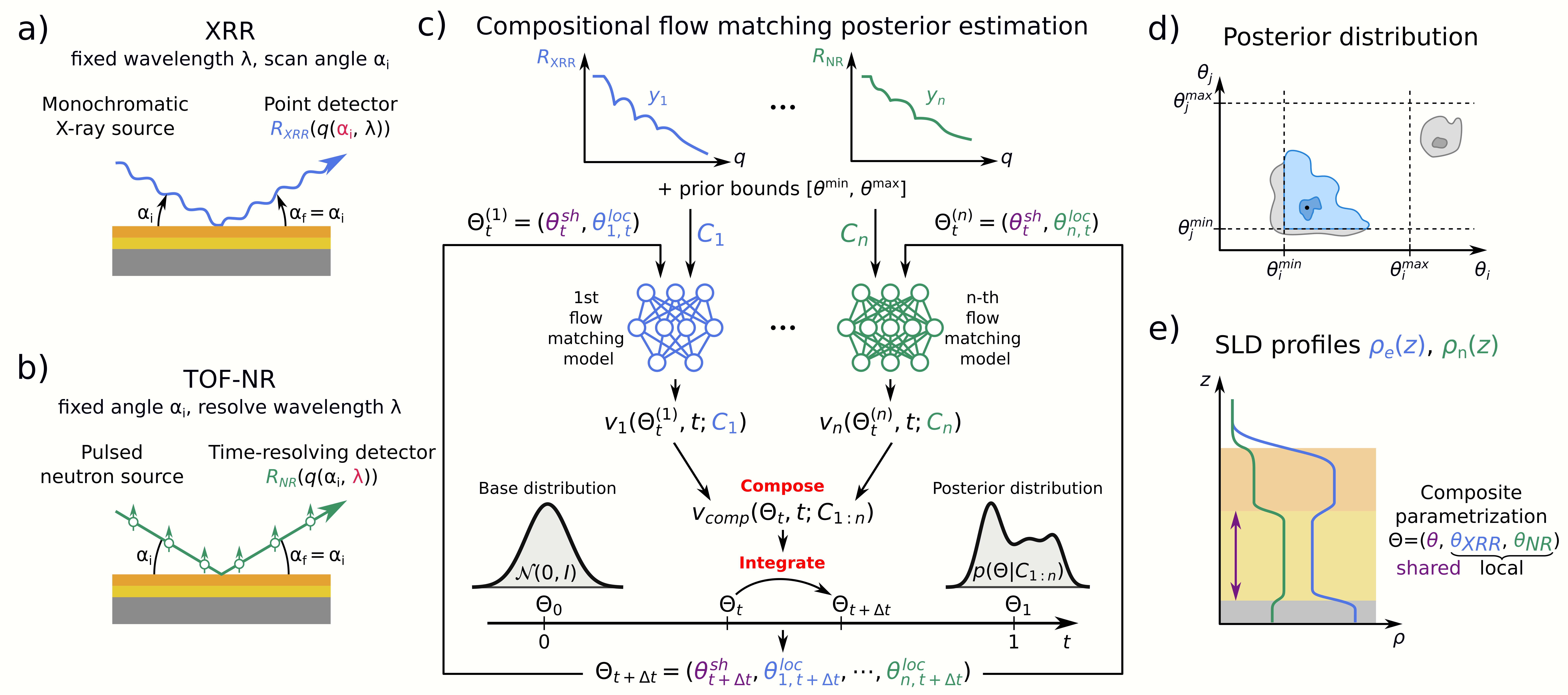}
  \caption{\textbf{Prior-amortized flow matching posterior estimation and
compositional co-refinement for reflectometry.}
(a) Schematic X-ray reflectometry (XRR) geometry. The reflectivity \(R_{\mathrm{XRR}}\) is measured by varying the incident angle \(\alpha_i\) at fixed wavelength \(\lambda\), giving access to \(R_{\mathrm{XRR}}(q(\alpha_i,\lambda))\).
(b) Schematic time-of-flight neutron reflectometry (TOF-NR) geometry. The incident angle is fixed, while the wavelength dependence is resolved from neutron time of flight, giving \(R_{\mathrm{NR}}(q(\alpha_i,\lambda))\).
(c) Compositional flow matching posterior estimation. Each observation \(y_j\), together with the corresponding prior bounds \(b_j\), defines a conditioning context \(C_j\). A single-observation flow matching model maps the relevant latent coordinates \(\Theta_t^{(j)}=(\theta_t^{\mathrm{sh}},\theta^{\mathrm{loc}}_{j,t})\) to a local vector-field contribution. The local contributions are embedded into a common shared/local parameter space, combined into a composed vector field, and integrated from a normal base distribution to obtain samples from the joint posterior over the global parameter vector \(\Theta=(\theta^{\mathrm{sh}},\theta^{\mathrm{loc}}_1,\ldots,\theta^{\mathrm{loc}}_n)\).
(d) Schematic two-dimensional posterior representation for parameters \(\theta_i\) and \(\theta_j\). The dashed lines indicate the user-supplied prior bounds, and the blue region denotes posterior density retained within the selected bounded parameter domain. Grey regions indicate density outside the selected bounds that is excluded by the bounded prior.
(e) Posterior parameter samples can be propagated through the physical sample parameterization to obtain depth-dependent SLD profiles. In multimodal XRR/NR co-refinement, shared structural parameters determine the common layer geometry, while probe-specific local parameters determine the electronic SLD profile \(\rho_e(z)\) and the nuclear SLD profile \(\rho_n(z)\).}
\label{fig:fmpe_overview}
\end{figure}

Figure~\ref{fig:fmpe_overview} summarizes the analysis strategy and its application to reflectometry. XRR and TOF-NR illustrate different acquisition modes for observations that each contain one reflectivity curve. In conventional XRR, the incident angle is varied at fixed wavelength, whereas in TOF-NR the incident angle is fixed and the wavelength dependence is resolved from neutron time of flight (Fig.~\ref{fig:fmpe_overview}a,b). Despite these experimental differences, each measurement is supplied to its corresponding posterior estimator together with the selected prior bounds. An observation can also comprise a fixed number of physically labeled channels. Non--spin-flip PNR observations contain two channels, while full spin-resolved PNR observations contain four channels. Because these channels have fixed physical labels and ordering, they are supplied jointly to the embedding network as channels of a single observation. This fixed multichannel treatment is distinct from compositional sampling, which combines separate observations, whether single- or multichannel, at inference time.

In practice, we use the trained posterior estimators in a two-stage inference procedure. First, raw flow matching sampling provides a fast amortized approximation to the posterior by integrating the learned single-observation or composed flow. Second, when likelihood-based refinement is desired, these samples initialize an ensemble MCMC sampler targeting the posterior defined by the explicit forward-model likelihood and the selected prior bounds. Unless stated otherwise, the main-text figures show posterior predictive simulations and profiles from this refined posterior, which we denote FMPE-MCMC. The Supplementary Information reports the corresponding raw FMPE results and corner-plot comparisons between the raw FMPE and FMPE-MCMC posteriors, providing a diagnostic of the amortized approximation before refinement.

The following sections first evaluate prior-amortized flow matching across single-observation XRR, NR, and PNR models. We then evaluate homogeneous and heterogeneous composition for multi-contrast NR and multimodal XRR/PNR co-refinement.

\clearpage

\subsection*{Single-curve XRR posterior inference for organic thin films}

We first consider single-curve XRR posterior inference for organic thin films. This example uses a PTCDI-C$_{3}$ film on a silicon substrate, while an analogous analysis of a C$_{60}$ film is provided in the Supplementary Information. Such films are representative of organic semiconductor systems, in which film thickness, density and interfacial morphology are important structural characteristics that can influence charge-transport behavior~\cite{Chesterfield2004PeryleneDiimides}. This tests the basic single-observation setting in which one trained prior-amortized model is applied to experimental curves using sample-specific prior bounds. For this application, we use a network trained for a standard one-layer slab parameterization. In addition to the structural parameters, the parameterization includes a \(q\)-shift offset to account for possible alignment errors and an intensity scale factor to account for normalization uncertainty \cite{Greco2022Neural}. The prior bounds supplied at inference time are wider for thicknesses and roughnesses and narrower for the SLDs, reflecting the fact that the film material is known. 

Figs.~\ref{fig:single_curve_examples}a and \ref{fig:single_curve_examples}b show the experimental PTCDI-C$_{3}$ curve together with posterior-predictive simulations from the FMPE-MCMC posterior and the corresponding SLD profiles. The posterior-predictive ensemble closely reproduces the experimental curve. The raw FMPE posterior already provides physically plausible samples, while the MCMC refinement targets the explicit likelihood and reduces residual amortization errors.

\clearpage

\subsection*{NR with time-of-flight data and structured brush parameterizations}

We next consider two neutron reflectometry examples that extend the prior-amortized framework beyond uniformly sampled and slab-style reflectivity curves. 

The first example uses time-of-flight NR data measured on the D17 reflectometer at the Institut Laue-Langevin and published by Higgins et al.~\cite{Higgins2024}. The sample is a two-layer PS:Bis-PCBM/Bis-PCBM structure on silicon, with air as the fronting medium. Bis-PCBM is a fullerene-derived electron acceptor used in organic photovoltaic blends, where polymer--acceptor miscibility and the resulting vertical composition and morphology can strongly influence device performance~\cite{Chen2014BisPCBMMiscibility}.

The second example uses NR data from a polystyrene (PS) polymer-brush structure originally reported by Hiotelis et al.~\cite{Hiotelis2008} and distributed with the \texttt{anaklasis} package~\cite{Koutsioubas2021}. Polymer brushes are widely used to regulate protein adsorption, cell attachment and bacterial adhesion in biosensing and biomedical coatings~\cite{Krishnamoorthy2014BiomedicalBrushes}. Their function depends on the depth-dependent polymer and solvent distributions at the interface, making the reconstructed brush height and solvent volume fraction profile physically relevant quantities. This example tests a structured physical parameterization in which the continuous brush profile is described by a small number of interpretable parameters and then discretized for reflectivity calculation.

For the time-of-flight example, we use a network trained for a standard two-layer slab parameterization with intensity scale and constant background as additional instrumental parameters. For the present inference workflow, TOF-NR provides a more demanding test than uniformly sampled angle-scan data because experimental curves can have non-uniform \(q\)-spacing and pointwise resolution information. The neural network is conditioned on a scalar linear resolution-smearing coefficient, whereas the MCMC refinement uses the pointwise resolution values available for the experimental data. This difference reflects the roles of the two stages: the scalar coefficient provides an instrument-agnostic approximation suitable for broad simulation-based training, while the likelihood-based refinement can use the instrument-specific resolution information directly. The example therefore tests whether the amortized model remains useful when the neural-network input uses a simplified resolution description but the final posterior refinement uses a more detailed experimental likelihood.

For the polymer-brush example, the model parameters describe the brush height, the polymer volume fraction at the grafting interface, the brush-profile exponent, the thickness and SLD of a thin layer between the fronting medium and the brush (reactive layer), and the relevant interfacial roughnesses. For the experimental structure considered here, the fronting medium is quartz, the backing medium is deuterated toluene, and the additional thin layer corresponds to water. The continuous brush profile and its discretization are defined in Methods. This case is representative of practical reflectometry models that are not simple stacks of independent slabs, but instead contain structured parameterizations derived from physical constraints or soft-matter theory.

\begin{figure}[t]
  \centering
  \includegraphics[width=1.0\linewidth]{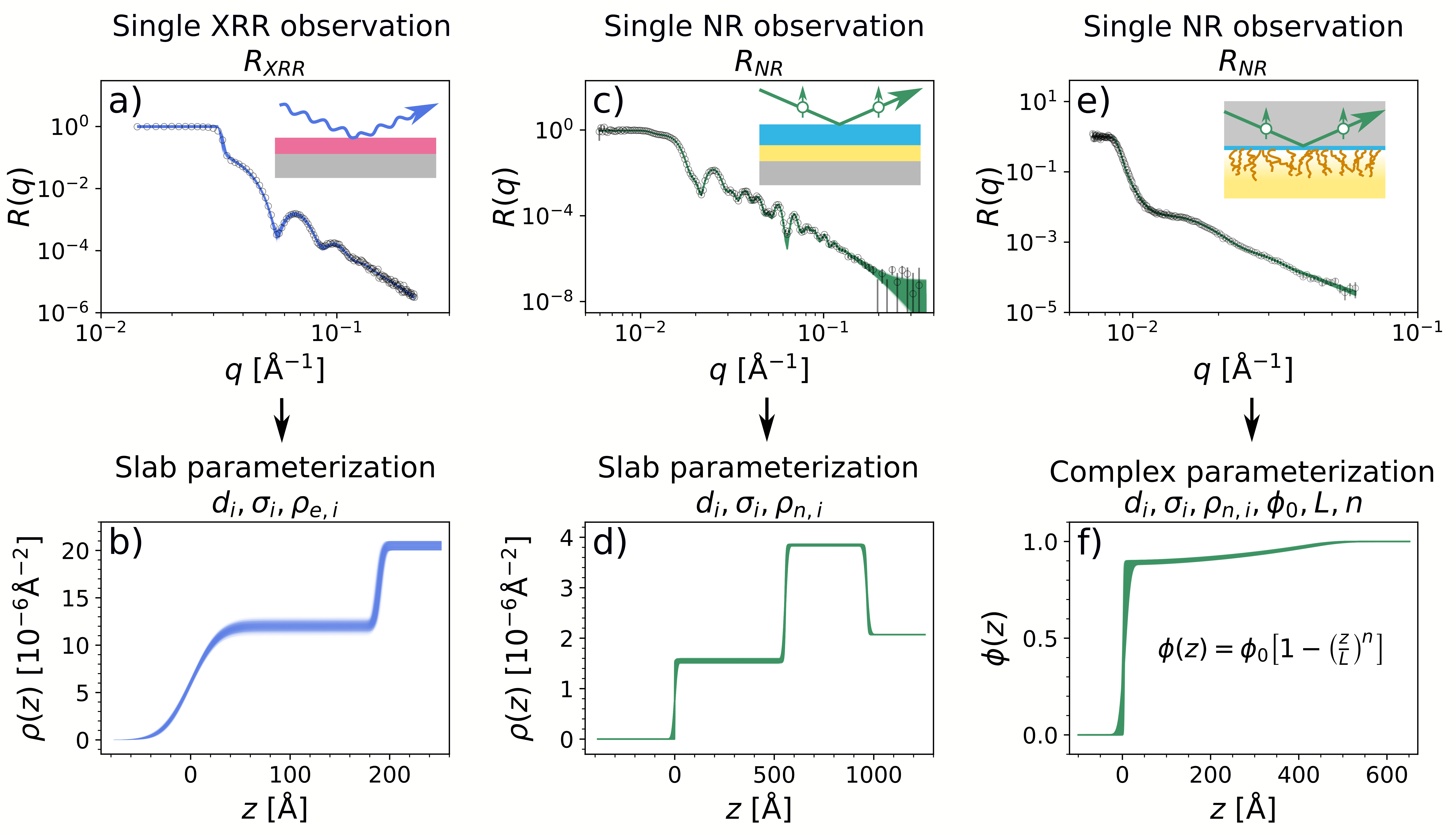}
  \caption{\textbf{Single-curve posterior inference for XRR, TOF-NR and polymer-brush NR data.}
  (a) Experimental XRR curve for a one-layer PTCDI-C$_{3}$ film on a Si substrate, shown together with posterior-predictive simulations generated from FMPE-MCMC posterior samples.
  (b) SLD profiles for the PTCDI-C$_{3}$ film reconstructed from the FMPE-MCMC posterior.
  (c) Experimental TOF-NR curve for a two-layer PS:Bis-PCBM/Bis-PCBM structure on Si, with data from Ref.~\cite{Higgins2024}, shown together with posterior-predictive simulations generated from FMPE-MCMC posterior samples.
  (d) SLD profiles reconstructed from the FMPE-MCMC posterior for the TOF-NR data.
  (e) Experimental NR curve for the polymer-brush structure, with data from Ref.~\cite{Koutsioubas2021}, shown together with posterior-predictive simulations generated from FMPE-MCMC posterior samples.
  (f) Solvent volume-fraction profiles reconstructed from the FMPE-MCMC posterior for the polymer-brush structure.}
  \label{fig:single_curve_examples}
\end{figure}

Fig.~\ref{fig:single_curve_examples}c shows the time-of-flight NR data together with posterior-predictive simulations from the FMPE-MCMC posterior, and Fig.~\ref{fig:single_curve_examples}d shows the corresponding SLD profiles. Fig.~\ref{fig:single_curve_examples}e shows the analogous result for the polymer-brush example, with the reconstructed solvent volume-fraction profiles shown in Fig.~\ref{fig:single_curve_examples}f. These results indicate that our posterior estimation method can be applied both to experimental TOF data and to structured model classes. The corresponding raw FMPE posterior-predictive curves and posterior corner plots are provided in the Supplementary Information.

\clearpage

\subsection*{Full spin-resolved PNR}

We use full spin-resolved PNR as a proof-of-concept for multichannel posterior estimation. Such measurements are particularly relevant for magnetic heterostructures with non-collinear or depth-dependent magnetization, including spin valves, exchange-spring systems and exchange-biased structures~\cite{Zabel2008PNRNanomagneticSystems}. In such systems, the non--spin-flip channels encode information about the nuclear SLD and the magnetization component parallel to the neutron-spin quantization axis, whereas the spin-flip channels provide information about transverse magnetization components. In contrast to standard XRR or NR, a full spin-resolved PNR observation therefore consists of four coupled reflectivity channels: two non--spin-flip channels and two spin-flip channels. In our framework these four curves are supplied jointly to the embedding network as a single multichannel observation, allowing the posterior estimator to learn correlations between nuclear, magnetic and angular parameters from the full spin-resolved signal. 

For this example, we trained a model with a simplified two-layer magnetic parameterization including layer thicknesses, nuclear SLDs, magnetic SLDs and magnetization angles. Training-data generation for this setting used a batched \(4\times4\) transfer-matrix implementation in PyTorch. Interfacial roughnesses are not included in this proof-of-concept model because of the limitations of the transfer-matrix treatment discussed in the Methods section. We apply the model to synthetic full spin-resolved PNR data emulating a Co/Fe bilayer on a silicon backing medium, with the Co magnetization oriented at \(45^\circ\) and the Fe magnetization oriented at \(120^\circ\) with respect to the applied magnetic field. Resolution smearing \(dq/q = 0.05\) was applied to the synthetic data.

Fig.~\ref{fig:PNR_and_NR_contrasts}a shows the four spin-resolved reflectivity channels together with posterior-predictive simulations from the FMPE-MCMC posterior. Figs.~\ref{fig:PNR_and_NR_contrasts}b and \ref{fig:PNR_and_NR_contrasts}c show the reconstructed nuclear and magnetic SLD profiles and the magnetization-angle profiles. The posterior concentrates around the parameters used to generate the synthetic data, showing that the flow matching posterior estimator can represent the coupled uncertainty in nuclear, magnetic and angular parameters in this simplified full spin-resolved PNR setting. This result should be viewed as a proof of concept for spin-resolved posterior inference rather than as a complete treatment of experimental full-PNR analysis with roughness and additional instrumental effects.

\clearpage

\subsection*{Compositional co-refinement of multiple-contrast NR data}

We next demonstrate compositional sampling for joint posterior inference from multiple-contrast NR data. Contrast variation is essential in many areas of soft-matter and interfacial science because it allows selected regions or molecular components of a sample to be highlighted without substantially changing the underlying structure. Different contrasts can be generated by varying the isotopic composition of the surrounding solvent or by selectively deuterating sample components such as polymers, lipids or proteins. Combining measurements at complementary contrasts improves sensitivity to different regions of the interfacial structure and helps constrain layer composition and solvent content~\cite{Cousin2018SoftMatterNR,Heinrich2020ContrastVariation}. In the present example, the contrast is varied through the composition of the backing solvent. This example uses a single neural network model trained on individual observations and composes the corresponding posterior information across three solvent contrasts at inference time. The model uses a two-layer slab parameterization with additional solvent volume-fraction parameters, which quantify penetration of the backing solvent into the layers.

The experimental data were published along with the \texttt{refnx} package~\cite{NelsonRefnx}. The sample consists of a silicon oxide layer and a polymer layer, with silicon as the fronting medium and a solvent as the backing medium. Three solvent contrasts are used: D$_2$O, H$_2$O and a silicon-oxide-matched mixture of D$_2$O and H$_2$O (in a 59:41 ratio), with an SLD of \(3.5\times10^{-6}\mathrm{\AA}^{-2}\). For compactness, this contrast is labelled CM in the figures. In the joint inference, the structural parameters are shared across all contrasts, whereas the backing-medium SLDs, intensity scale factors and backgrounds are treated as observation-specific parameters. This shared/local split is the key reflectometry-specific extension of the compositional setting, because different measurements of the same sample generally share the structure but not all contrast or instrumental parameters.

\begin{figure}[h!]
  \centering
  \includegraphics[width=1.0\linewidth]{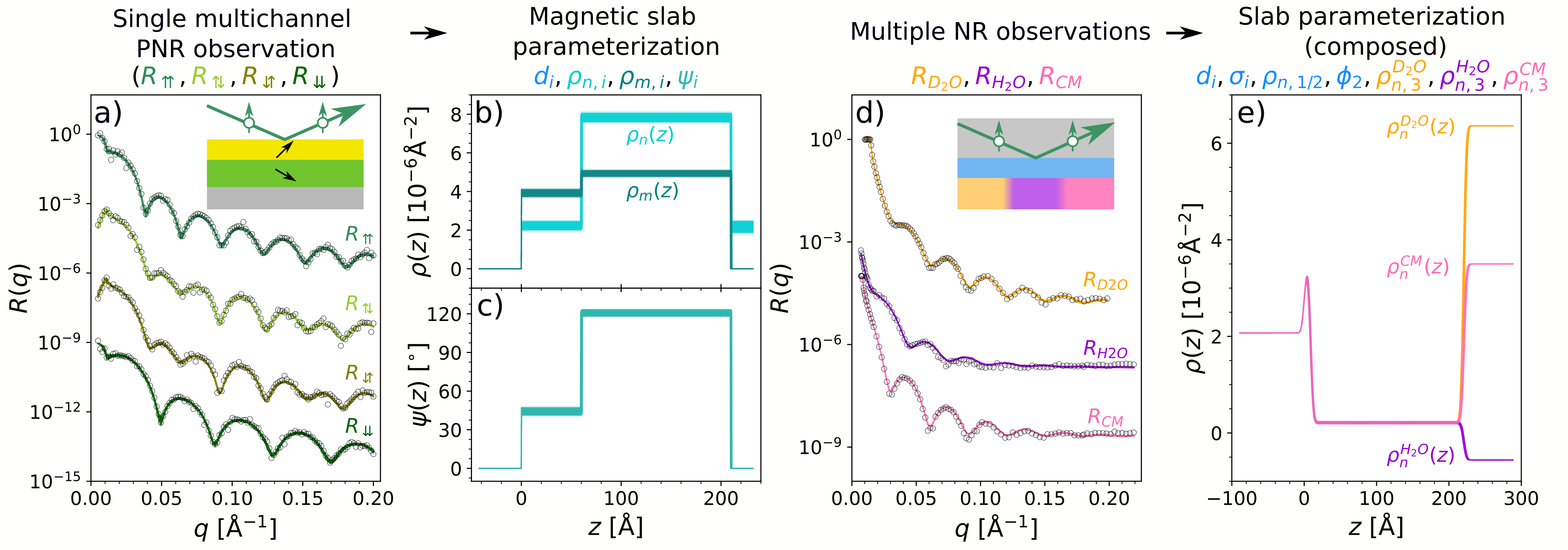}
  \caption{\textbf{Multichannel posterior inference for full spin-resolved PNR and compositional co-refinement of multiple-contrast NR data.}
  (a) Synthetic full spin-resolved PNR data, shown together with posterior-predictive simulations generated from FMPE-MCMC posterior samples. The curves corresponding to different spin channels are shifted vertically for clarity.
  (b) Nuclear and magnetic SLD profiles reconstructed from the FMPE-MCMC posterior.
  (c) Magnetization-angle profiles reconstructed from the FMPE-MCMC posterior.
  (d) Experimental NR curves corresponding to three solvent contrasts, D$_2$O, H$_2$O and CM, with data from Ref.~\cite{NelsonRefnx}, shown together with posterior-predictive simulations generated from FMPE-MCMC posterior samples. The curves are shifted vertically for clarity.
  (e) SLD profiles for the three solvent contrasts reconstructed from the FMPE-MCMC posterior.}
  \label{fig:PNR_and_NR_contrasts}
\end{figure}

Fig.~\ref{fig:PNR_and_NR_contrasts}d shows the three measured NR curves together with posterior-predictive simulations from the FMPE-MCMC posterior, and Fig.~\ref{fig:PNR_and_NR_contrasts}e shows the corresponding contrast-dependent SLD profiles. The joint posterior enforces a single structural description across all contrasts while allowing each contrast to retain its own solvent and instrumental parameters. This result shows that a model trained only on single observations can be reused for multi-contrast co-refinement, avoiding the need to train a separate network for each possible number or combination of contrasts. The corresponding raw-FMPE posterior-predictive curves and posterior corner plots are provided in the Supplementary Information.

\clearpage

\subsection*{Multimodal co-refinement of XRR and PNR data}

Finally, we consider a multimodal co-refinement setting in which different experimental probes are analyzed jointly using different trained posterior estimation models. This setting is important for reflectometry because XRR and (P)NR are sensitive to different material contrasts while sharing the same underlying layer geometry. We therefore compose an XRR model and a non--spin-flip PNR model at inference time, assigning layer thicknesses and interfacial roughnesses as shared parameters and retaining probe-specific SLD and instrumental parameters as local parameters.

The XRR model includes both the real and imaginary parts of the electronic SLD, because absorption effects cannot be neglected for the high-density materials considered here. The PNR model includes nuclear and magnetic SLDs and a channel-mixing coefficient \(\alpha\) to account for imperfect polarization. Both models include intensity scale and background parameters. We apply the approach to XRR and PNR measurements of a Pt/Fe structure on MgO, published along with the \texttt{anaklasis} package~\cite{Koutsioubas2021}. Such heterostructures are representative heavy-metal/ferromagnet systems investigated in connection with phenomena including interfacial magnetic anisotropy and spin--orbit torques~\cite{Eimer2023PtCoSpintronics,Hayashi2021PtSpinOrbitTorques}. For the main inference, the SLD parameters are fixed to their theoretical values, following the corresponding classical analysis; an additional inference with narrow SLD prior bounds is provided in the Supplementary Information.

Fig.~\ref{fig:XRR_and_PNR}a shows the measured XRR and PNR curves for the Pt/Fe structure together with posterior-predictive simulations from the FMPE-MCMC posterior. Fig.~\ref{fig:XRR_and_PNR}b shows the reconstructed real and imaginary electronic SLD profiles, and Fig.~\ref{fig:XRR_and_PNR}c shows the corresponding nuclear and magnetic SLD profiles. The results illustrate heterogeneous compositional sampling across different trained models and different experimental modalities. In this setting, XRR and PNR jointly constrain the shared structural parameters while preserving probe-specific information.

\begin{figure}[h!]
  \centering
  \includegraphics[width=1.0\linewidth]{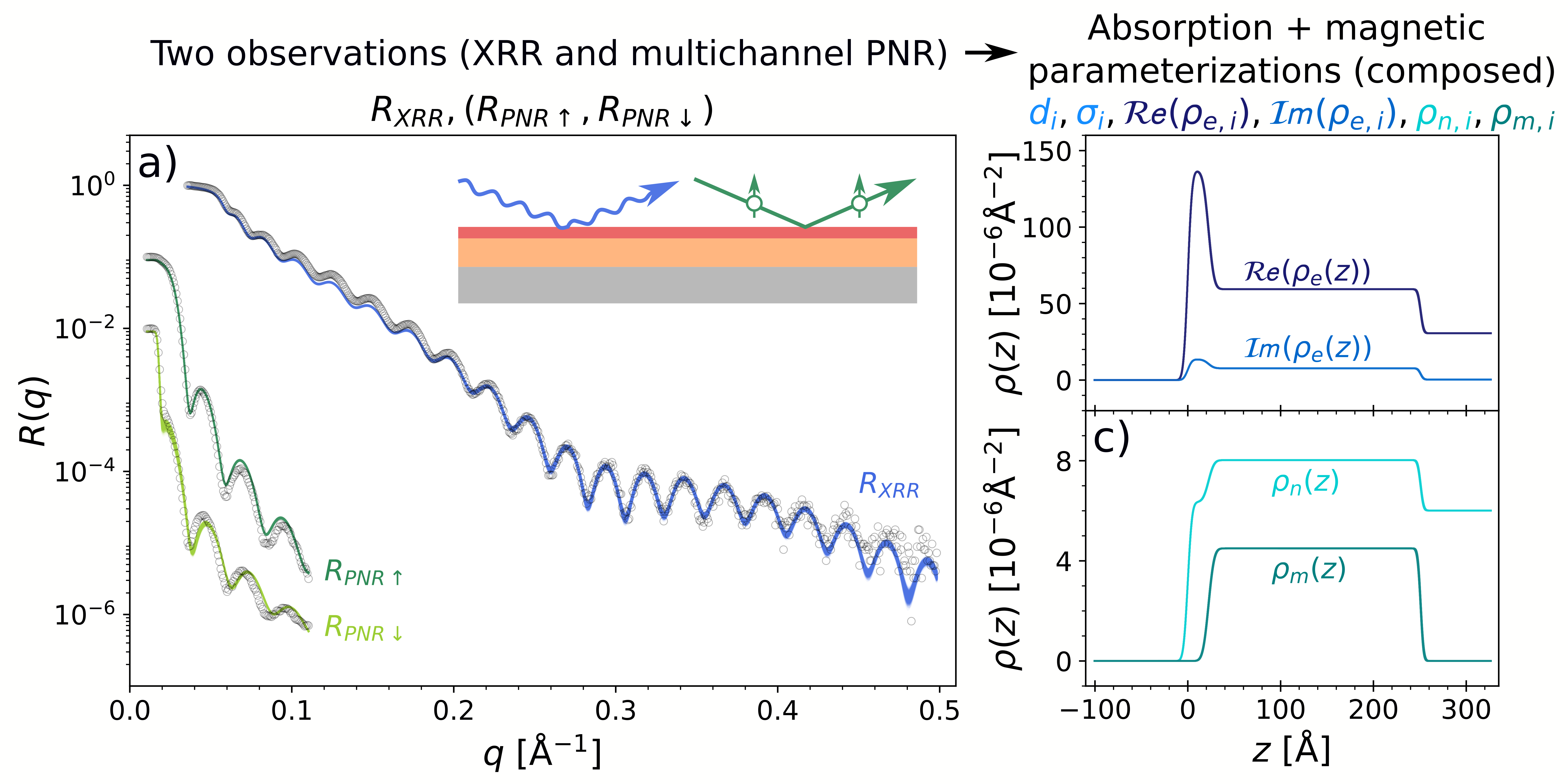}
  \caption{\textbf{Multimodal compositional co-refinement of XRR and PNR data.}
  (a) Experimental XRR and PNR data for the Pt/Fe structure, with data from Ref.~\cite{Koutsioubas2021}, shown together with posterior-predictive simulations generated from FMPE-MCMC posterior samples.
  (b) Reconstructed real and imaginary electronic SLD profiles.
  (c) Reconstructed nuclear and magnetic SLD profiles.}
  \label{fig:XRR_and_PNR}
\end{figure}

\clearpage

\section*{Discussion}

Probabilistic inversion of scattering data requires methods that can represent uncertainty and combine information from complementary measurements. Here, we introduced prior-amortized flow matching posterior estimation for X-ray and neutron reflectometry together with an inference-time compositional strategy for co-refinement. Compared with point-estimation approaches, the method 
directly returns posterior samples over physical and instrumental parameters, which is particularly valuable for inverse problems in which different parameter combinations can produce similar reflectivity curves. Flow matching provides a continuous-time generative formulation based on a learned vector field. For the Gaussian probability path used here, this vector field also provides access to a score representation, enabling posterior information from multiple observations to be composed at inference time. More generally, the trained FMPE models can cover broad parameter and prior-bound ranges and can be applied to different experimental datasets within the support of their training distributions. Depending on the measurement setting, an FMPE model may process a single reflectivity curve, a fixed set of physically labeled channels within one observation, as in PNR, or contribute to the compositional analysis of separate observations, as in isotope-contrast series and multimodal XRR/NR measurements.

A central contribution is the compositional sampling strategy for reflectometry co-refinement. Building on previous compositional SBI formulations, we derive a full-covariance composition rule for estimators defined on a common block of shared parameters together with observation-specific local parameter blocks. This structure is particularly relevant to reflectometry, where some parameters are shared because they describe the same sample geometry, whereas others are local because they describe contrast conditions, instrumental effects or probe-specific scattering properties. The shared--local formulation makes it possible to combine compatible single-observation posterior estimators for multi-contrast NR and heterogeneous XRR/(P)NR co-refinement. Beyond the reflectometry cases considered here, the approach may also be useful for probabilistic analysis of other multi-measurement experiments in materials science. Many experimental workflows combine observations that probe a common latent sample state while retaining measurement-specific contrast or instrumental parameters. Examples include measurements collected at different contrasts, incident energies, instruments or experimental modalities, as well as temperature- or time-dependent series in which the shared and evolving quantities can be specified explicitly. In such settings, a single-observation posterior estimator could, in principle, be reused as a modular component of joint inference. This modularity separates the reusable task of developing robust single-observation models from the experiment-specific task of composing the measurements available for a particular sample.

The examples show that the same framework can be applied to several experimentally relevant settings: single-curve XRR, time-of-flight NR, structured parameterizations such as polymer brushes, PNR and multi-measurement co-refinement. In many cases, raw FMPE samples already provide useful posterior approximations, whereas MCMC refinement initialized from those samples targets the posterior defined by the explicit forward-model likelihood and prior. This two-stage procedure is useful in practice because the amortized model identifies plausible regions of parameter space, while the refinement step reduces residual approximation errors and incorporates experimental likelihood details that may not be fully represented during neural network training.

Despite these promising results, the framework also has several limitations. Posterior samples from the amortized model are approximate, and their accuracy depends on the training ranges, the parameter dimensionality and the supplied prior bounds. For full spin-resolved PNR, the present demonstration uses synthetic data and a simplified parameterization without interfacial roughness. In addition, flow matching inference requires numerical integration of an ODE, which can make raw sampling slower than direct sampling from many discrete normalizing-flow models. Compositional accuracy additionally depends on the quality of the individual score estimates and covariance approximations, as well as on compatible definitions of the shared parameters, priors and probability paths across the contributing estimators. These limitations suggest several directions for further development. Approximation errors may be reduced through advances in flow matching objectives, probability-path design and training strategies~\cite{Lipman2024FlowMatchingGuide}, or through post-training refinement methods that incorporate feedback from the forward simulator~\cite{Holzschuh2024SimulatorFeedback}. Moreover, recent average-velocity and caching-based approaches provide potential routes to accelerating flow-based sampling~\cite{Geng2025MeanFlow,Gao2026MeanCache}. Future work could improve the compositional sampler through more accurate intermediate time covariance approximations and extensions to more general parameter-sharing patterns.

Overall, the results show that flow matching provides a flexible framework for probabilistic reflectometry inversion and that compositional sampling can extend single-observation posterior estimators to modular co-refinement. Although demonstrated here for reflectometry, the shared--local construction could be transferred to other multi-measurement scattering problems in which several observations constrain a common physical structure while retaining contrast-, probe- or instrument-specific parameters. This capability is relevant to high-throughput and multi-probe X-ray and neutron studies, where scalable uncertainty-aware analysis is increasingly important for automated experimental workflows.

\clearpage

\section*{Methods}
\label{sec:methods}

\subsection*{Bayesian reflectometry and prior amortization}

Let $\theta$ denote the vector of physical and instrumental parameters of a reflectometry model and let $y$ denote measured reflectivity data. Depending on the parameterization, $\theta$ may include layer thicknesses $d$, interfacial roughnesses $\sigma$, electronic, nuclear or magnetic SLDs $\rho$, magnetization angles $\psi$, solvent volume fractions $\phi$, and instrumental parameters such as intensity scale factors ($r_{\mathrm{sc}}$), $q$-offsets ($q_{\mathrm{sh}}$), or constant backgrounds.

We denote by $b$ the prior-conditioning variables supplied at inference time. In this work, these are the lower and upper bounds $[b_{\min},b_{\max}]$ for each parameter. These bounds encode experiment-specific prior knowledge and define a conditional prior $p(\theta\mid b)$.

We use $\eta$ to denote optional known experimental variables that are supplied to the neural network. For the NR models considered here, $\eta$ contains the linear smearing coefficient $dq/q$ and for the XRR models $\eta$ is empty.
The posterior of interest is then
\begin{equation}
p(\theta\mid y,b,\eta)
\propto
p(y\mid\theta,\eta)\,p(\theta\mid b).
\label{eq:conditional_reflectometry_posterior}
\end{equation}

In the prior-amortized setting, the neural network is trained across a broad distribution of prior bounds and observations rather than for a single fixed prior. 

% This setup follows the PANPE framework~\cite{vladimir_panpe} and is particularly natural in reflectometry, where realistic prior intervals vary from sample to sample.

Throughout the paper, we write
\begin{equation}
C=(y,b,\eta)
\label{eq:conditioning_context}
\end{equation}
for the complete conditioning context supplied to the neural posterior estimator. When $\eta$ is empty, this reduces to $C=(y,b)$. The goal of the generative model is to produce approximate posterior samples
\begin{equation}
\theta\sim p(\theta\mid C).
\end{equation}

\subsection*{Flow matching posterior estimation}

Flow matching \cite{Wildberger_NeurIPS_FlowMatchingSBI} parameterizes a time-dependent conditional vector field
\(
v(\theta,t;C)
\)
that transports noise sampled from a simple base distribution \(p_0(\theta)\) at the \textit{path time} \(t=0\) to the target posterior \(p(\theta \mid C)\) at \(t=1\), using a neural network. The associated probability path \(\{p_t(\theta\mid C)\}_{t\in[0,1]}\) satisfies the \textit{continuity equation}
\begin{equation}
\partial_t p_t(\theta \mid C) + \nabla_\theta \cdot \bigl( p_t(\theta \mid C)\, v(\theta,t;C) \bigr) = 0.
\end{equation} 

A direct regression objective based on the marginal vector field of \(p_t(\theta\mid C)\) is generally intractable, since this is not available in closed form. Flow matching circumvents this difficulty by introducing a family of \emph{conditional probability paths}, indexed by an auxiliary variable \(z\), together with corresponding conditional target vector fields \(u_t(\theta \mid z,C)\). The resulting conditional flow matching objective is
\begin{equation}
\mathcal{L}_{\mathrm{CFM}}
=
\mathbb{E}_{t,z,\theta_t}
\left[
\left\|
v(\theta_t,t;C) - u_t(\theta_t \mid z,C)
\right\|^2
\right],
\end{equation}
where \(\theta_t \sim p_t(\theta \mid z,C)\). Lipman et al. \cite{lipman2023flow} showed that this objective provides an unbiased gradient estimator for the corresponding marginal flow matching objective, while subsequent work generalized the framework and emphasized the influence of the chosen path family on training and sampling behavior \cite{tong2024improving,albergo2025stochasticinterpolantsunifyingframework}.

For posterior estimation, the target endpoint distribution is \(p(\theta\mid C)\). Accordingly, the neural network is trained as a conditional generative model, with the context \(C\) supplied as additional input.

In the two-sided path construction \cite{tong2024improving}, the auxiliary variable is
\(
z=(\theta_0,\theta_1)
\),
where
\(
\theta_0 \sim p_0
\)
and
\(
\theta_1 \sim p(\theta \mid C)
\).
For the straight-line path, the interpolation is
\begin{equation}
\theta_t = (1-t)\theta_0 + t\theta_1,
\end{equation}
and the corresponding conditional target vector is
\begin{equation}
u_t(\theta_t \mid \theta_0,\theta_1,C) = \theta_1 - \theta_0.
\end{equation}
This straight-path construction is the simplest and most widely used choice in rectified-flow and two-sided conditional flow matching variants \cite{tong2024improving,liu2023flow}. 

Flow matching is not intrinsically restricted to Gaussian base distributions
or to diffusion probability paths. In this work, however, we use a base noise
variable
\(
\epsilon\sim\mathcal{N}(0,I)
\),
for which the chosen interpolation defines a Gaussian conditional probability
path. Such paths admit both vector field and score representations, providing
a connection between deterministic flow and score-based formulations
\cite{albergo2025stochasticinterpolantsunifyingframework}. The path can be written
as
\begin{equation}
\theta_t
=
m_t\theta_1+\sigma_t\epsilon.
\label{eq:gaussian_probability_path}
\end{equation}
In our implementation, we use
\begin{equation}
m_t=t,
\qquad
\sigma_t=1-t+\sigma_{\min},
\qquad
\sigma_{\min}=10^{-3},
\label{eq:fm_path_coefficients}
\end{equation}
where the small endpoint regularization prevents numerical singularities near
\(t=1\).

For a Gaussian conditional probability path of the form in
Eq.~\eqref{eq:gaussian_probability_path}, the vector field
\(v(\theta_t,t;C)\) and the score
\begin{equation}
s(\theta_t,t;C)
:=
\nabla_{\theta_t}\log p_t(\theta_t\mid C)
\end{equation}
are related by~\cite{ma2024sitexploringflowdiffusionbased}
\begin{equation}
s(\theta_t,t;C)
=
\frac{
m_t v(\theta_t,t;C)-\dot m_t\theta_t
}{
\sigma_t
\left(
\dot m_t\sigma_t-m_t\dot\sigma_t
\right)
}.
\label{eq:general_score_from_flow}
\end{equation}
For the coefficients in Eq.~\eqref{eq:fm_path_coefficients},
\(\dot m_t=1\) and \(\dot\sigma_t=-1\), such that
\begin{equation}
s(\theta_t,t;C)
=
\frac{
t\,v(\theta_t,t;C)-\theta_t
}{
(1-t+\sigma_{\min})(1+\sigma_{\min})
}.
\label{eq:score_from_flow}
\end{equation}

Related connections between deterministic flows and score-based
stochastic sampling are discussed in
Ref.~\cite{singh2024stochasticsamplingdeterministicflow}.

Once the model has been trained, approximate posterior samples are obtained by integrating the neural ordinary differential equation (ODE)~\cite{NEURIPS2018_69386f6b}
\begin{equation}
\frac{d\theta_t}{dt} = v(\theta_t,t;C),
\qquad
\theta_0 \sim p_0,
\end{equation}
from \(t=0\) to \(t=1\), using either a fixed number of integration steps (e.g. Euler solver) or an adaptive solver such as Dormand--Prince (\texttt{dopri5}) from the \texttt{torchdiffeq} package, which we use as the default solver.

\subsection*{Neural network architecture}

\begin{figure}[t]
  \centering
  \includegraphics[width=1.0\linewidth]{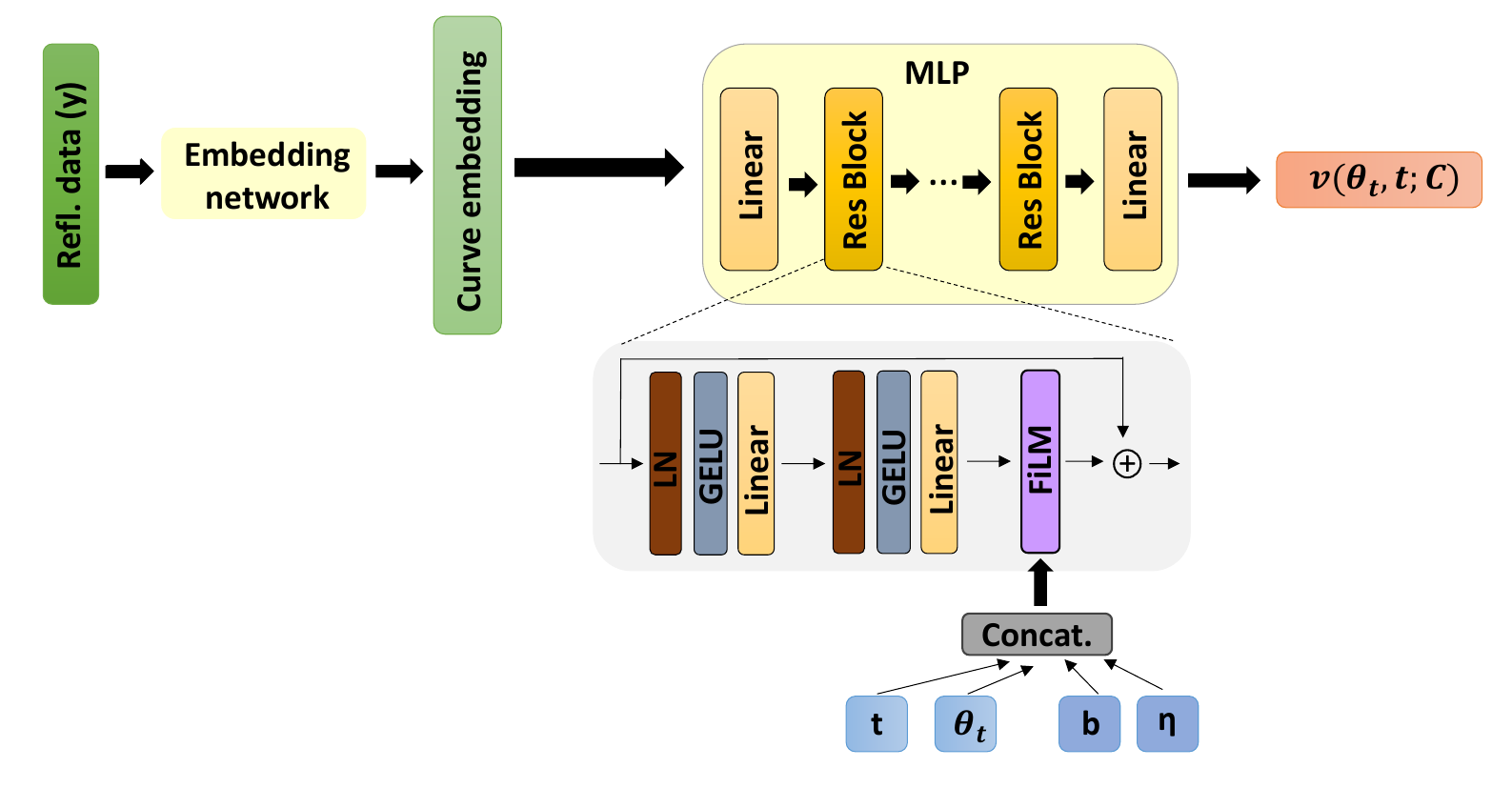}
  \caption{\textbf{Neural network architecture.} The reflectivity data is first mapped to a fixed-dimensional representation by an embedding network. This embedding is processed by a residual MLP with GELU activations and layer-normalized residual blocks. Conditioning is applied through FiLM layers using $t$, $\theta_t$, the prior bounds $b$, and optional auxiliary variables $\eta$. For the neutron-reflectometry models considered here, $\eta$ includes the linear smearing coefficient $dq/q$. The network outputs the conditional vector field $v(\theta_t,t;C)$.}
  \label{fig:neural_network_architecture_fm}
\end{figure}

The conditional vector field $v(\theta_t,t;C)$ is represented by a neural network consisting of two main components: an embedding network for the reflectometry observation and a residual multilayer perceptron (MLP) that outputs the vector field. The neural network architecture is illustrated in Fig.~\ref{fig:neural_network_architecture_fm}.

The observation $y$ is represented as multichannel one-dimensional data. Standard XRR and NR observations contain one reflectivity channel, non--spin-flip PNR observations contain two reflectivity channels, and full spin-resolved PNR observations contain four reflectivity channels. The corresponding $q$ values are supplied as an additional input channel. These inputs are mapped to a fixed-dimensional representation by a one-dimensional convolution-based embedding network adapted from previous work in the field~\cite{Munteanu2024,vladimir_panpe}.

The embedded representation is then processed by a residual MLP with $8$ residual blocks of width $1024$, GELU activations, and layer normalization. Conditioning is applied through feature-wise linear modulation (FiLM)~\cite{perez2018film}: the reflectivity embedding is provided as the main input to the residual MLP, while $t$, $\theta_t$, the prior bounds $b$, and any optional auxiliary conditioning variables $\eta$ are injected through FiLM layers in each residual block. For the neutron reflectometry applications considered here, $\eta$ includes the linear smearing coefficient $dq/q$. The final linear layer is initialized with zero weights.

\clearpage

\subsection*{Compositional sampling for multiple observations}

Many reflectometry studies collect several measurements of the same underlying
sample, for example NR data recorded under different solvent contrasts or
complementary measurements obtained from different probes. The goal is to
perform joint posterior inference across all observations while using models
trained on single-observation tasks. Compositional posterior inference provides
a way to do this. In SBI, composition rules of this type were introduced for
score-based models by Ref.~\cite{geffner2023compositional} and were later
reformulated in a more stable framework by
Ref.~\cite{linhart2026diffusionposteriorsamplingsimulationbased}. Our
implementation was informed by the public \texttt{sbi}
ecosystem~\cite{Boelts2025_sbi}, but the formulation used here adapts the
tall-data construction, together with the full-covariance \textsc{Gauss}
composition of
Ref.~\cite{linhart2026diffusionposteriorsamplingsimulationbased}, to flow
matching posterior estimators and to the shared/local parameter structure
required for reflectometry co-refinement.

We start with the standard tall-data identity. If observations
\(y_{1:n}=(y_1,\ldots,y_n)\) are conditionally independent given a common
parameter vector \(\theta\), and if each single-observation posterior is
defined with respect to the same prior \(p(\theta\mid b)\), then
\begin{equation}
p(\theta\mid y_{1:n},b)
\propto
p(\theta\mid b)^{1-n}
\prod_{j=1}^{n}
p(\theta\mid y_j,b).
\label{eq:joint_posterior_factorization}
\end{equation}
The factor \(p(\theta\mid b)^{1-n}\) corrects for the repeated prior
contribution: multiplying \(n\) single-observation posteriors counts the prior
\(n\) times, whereas the joint posterior should contain it only once.

Reflectometry co-refinement requires an extension of this identity because not
all parameters are common to all measurements. Some parameters, such as layer
thicknesses and interfacial roughnesses, may be shared across observations,
whereas others, such as contrast-dependent SLDs, scale factors, backgrounds,
or probe-specific quantities, may be local to individual observations. We
therefore write the global parameter vector as
\begin{equation}
\Theta
=
\left(
\theta^{\mathrm{sh}},
\theta^{\mathrm{loc}}_1,
\ldots,
\theta^{\mathrm{loc}}_n
\right),
\label{eq:global_parameter_vector}
\end{equation}
where \(\theta^{\mathrm{sh}}\) denotes the shared block and
\(\theta^{\mathrm{loc}}_j\) denotes the block local to observation \(j\).

For each observation, the complete context supplied to the single-observation
posterior estimator is
\begin{equation}
C_j=(y_j,b_j,\eta_j),
\qquad
b_j=
\left(
b^{\mathrm{sh}},
b_j^{\mathrm{loc}}
\right),
\label{eq:single_observation_context_composition}
\end{equation}
where \(y_j\) is the measured reflectometry observation,
\(b^{\mathrm{sh}}\) contains the prior bounds for the shared parameters,
\(b_j^{\mathrm{loc}}\) contains the prior bounds for the local parameters of
observation \(j\), and \(\eta_j\) denotes optional known observation-specific
conditioning variables. We write \(C_{1:n}=\{C_1,\ldots,C_n\}\) for the full
set of contexts used in the joint analysis.

We assume conditional independence of the observations given the global
parameter vector,
\begin{equation}
p\!\left(
y_{1:n}\mid\Theta,\eta_{1:n}
\right)
=
\prod_{j=1}^{n}
p_j\!\left(
y_j
\mid
\theta^{\mathrm{sh}},
\theta_j^{\mathrm{loc}},
\eta_j
\right),
\label{eq:conditional_independence}
\end{equation}
where the subscript on \(p_j\) emphasizes that the likelihood functions may
differ between observations, as when different probes, instruments, or
forward models are combined, and each likelihood depends only on the
parameters used by its observation. Together with the factorized global prior
\begin{equation}
p\!\left(
\Theta
\mid
b^{\mathrm{sh}},
b^{\mathrm{loc}}_{1:n}
\right)
=
p\!\left(
\theta^{\mathrm{sh}}
\mid
b^{\mathrm{sh}}
\right)
\prod_{j=1}^{n}
p\!\left(
\theta_j^{\mathrm{loc}}
\mid
b_j^{\mathrm{loc}}
\right),
\label{eq:factorized_global_prior}
\end{equation}
the target joint posterior can be written as
\begin{equation}
p(\Theta\mid C_{1:n})
\propto
p\!\left(
\theta^{\mathrm{sh}}
\mid
b^{\mathrm{sh}}
\right)^{1-n}
\prod_{j=1}^{n}
p_j\!\left(
\theta^{\mathrm{sh}},
\theta_j^{\mathrm{loc}}
\mid
C_j
\right).
\label{eq:joint_posterior_shared_local}
\end{equation}
Equation~\eqref{eq:joint_posterior_shared_local} is the shared/local analogue
of Eq.~\eqref{eq:joint_posterior_factorization}. The correction is applied
only to the shared prior because the shared prior appears once in every
single-observation posterior factor. In contrast, each local block
\(\theta_j^{\mathrm{loc}}\) appears in only one factor and therefore requires
no repeated-prior correction. Equation~\eqref{eq:joint_posterior_shared_local}
places no factorization requirement on the single-observation posteriors
themselves: the likelihood generally correlates the shared and local
parameters of each observation, and the composition below retains these
correlations.

The composition is performed in the parameter coordinates used by the neural
posterior estimators. For the shared parameters, all estimators entering the
composition must use the same coordinate representation. At path time \(t\),
the global latent state is written as
\begin{equation}
\Theta_t
=
\left(
\theta_t^{\mathrm{sh}},
\theta_{1,t}^{\mathrm{loc}},
\ldots,
\theta_{n,t}^{\mathrm{loc}}
\right),
\qquad
\Theta_t^{(j)}
=
P_j\Theta_t
=
\left(
\theta_t^{\mathrm{sh}},
\theta_{j,t}^{\mathrm{loc}}
\right),
\label{eq:global_and_local_latent_states}
\end{equation}
where the selection matrix \(P_j\) extracts the coordinates used by the
single-observation model for observation \(j\), its transpose
\(P_j^{\mathsf T}\) embeds observation-\(j\) quantities into the global
coordinates by zero padding, and \(P_{\mathrm{sh}}\) analogously selects the
shared block. The corresponding single-observation flow matching model
predicts the learned vector field
\begin{equation}
v_{j}\!\left(
\Theta_t^{(j)},t;C_j
\right).
\label{eq:local_vector_field}
\end{equation}

For compactness, we denote the associated learned score estimate by
\begin{equation}
s_{j,t}
\equiv
s_{j}\!\left(
\Theta_t^{(j)},t;C_j
\right).
\end{equation}
Using the probability-path coefficients
\(m_t=t\) and
\(\sigma_t=1-t+\sigma_{\min}\), this score is obtained from
Eq.~\eqref{eq:score_from_flow} as
\begin{equation}
s_{j,t}
\approx
\frac{
t\,v_{j}\!\left(
\Theta_t^{(j)},t;C_j
\right)
-
\Theta_t^{(j)}
}{
(1-t+\sigma_{\min})(1+\sigma_{\min})
}.
\label{eq:local_score_from_flow}
\end{equation}
This conversion is specific to the flow matching formulation used here. The
subsequent composition follows the \textsc{Gauss} precision-weighted score
composition of Ref.~\cite{linhart2026diffusionposteriorsamplingsimulationbased}: each
single-observation contribution is treated using a Gaussian approximation to
its denoising distribution at path time \(t\), and information from
independent measurements is combined by adding precision matrices. 

Each clean single-observation posterior is approximated by a Gaussian in its
local parameter coordinates,
\begin{equation}
p_j\!\left(
\Theta^{(j)}
\mid
C_j
\right)
\approx
\mathcal N(\mu_j,\Sigma_j),
\qquad
\Lambda_j=\Sigma_j^{-1}.
\label{eq:gaussian_endpoint_models}
\end{equation}
For the Gaussian probability path, the corresponding denoising precision at
path time \(t\) is
\begin{equation}
\Lambda_{j,t}
=
\frac{m_t^2}{\sigma_t^2}I
+
\Lambda_{j}.
\label{eq:factor_denoising_precision}
\end{equation}
This is a full matrix. Its shared--local off-diagonal blocks retain posterior
correlations that would be lost under a diagonal approximation.

In practice, \(\Lambda_{j}\) is estimated per context from the amortized
model itself: we draw a batch of raw FMPE samples from
\(p_j(\cdot\mid C_j)\), compute their empirical covariance
\(\widehat\Sigma_{j}\), and set
\(\Lambda_{j}=\widehat\Sigma_{j}^{-1}\). 

The prior enters the composition through two distinct quantities: the score
\(s_t^{\mathrm{pr}}\) of the perturbed shared prior and the precision
\(\Lambda_t^{\mathrm{pr}}\) of its denoising distribution. Define the
perturbed shared-prior density by
\begin{equation}
p_t^{\mathrm{pr}}\!\left(
\theta_t^{\mathrm{sh}}
\right)
=
\int
k_t\!\left(
\theta_t^{\mathrm{sh}}
\mid
\theta^{\mathrm{sh}}
\right)
p\!\left(
\theta^{\mathrm{sh}}
\mid
b^{\mathrm{sh}}
\right)
\,\mathrm d\theta^{\mathrm{sh}},
\label{eq:perturbed_shared_prior}
\end{equation}
where \(k_t\) is the Gaussian probability-path kernel. Its score is
\begin{equation}
s_t^{\mathrm{pr}}\!\left(
\theta_t^{\mathrm{sh}}
\right)
=
\nabla_{\theta_t^{\mathrm{sh}}}
\log
p_t^{\mathrm{pr}}\!\left(
\theta_t^{\mathrm{sh}}
\right).
\label{eq:diffused_prior_score_general}
\end{equation}
Given this score, the precision of the corresponding denoising distribution
follows from the second-order Tweedie identity,
\begin{equation}
\Lambda_t^{\mathrm{pr}}\!\left(
\theta_t^{\mathrm{sh}}
\right)
=
\frac{m_t^2}{\sigma_t^2}
\left[
I
+
\sigma_t^2
\nabla_{\theta_t^{\mathrm{sh}}}
s_t^{\mathrm{pr}}\!\left(
\theta_t^{\mathrm{sh}}
\right)
\right]^{-1}.
\label{eq:prior_precision_tweedie}
\end{equation}

Let
\begin{equation}
s_{\mathrm{comp},t}
\equiv
s_{\mathrm{comp}}\!\left(
\Theta_t,t;C_{1:n}
\right).
\end{equation}
The composed score is defined as the solution of the linear system
\begin{equation}
\Lambda_{\mathrm{comp},t}
s_{\mathrm{comp},t}
=
(1-n)
P_{\mathrm{sh}}^{\mathsf T}
\Lambda_t^{\mathrm{pr}}
s_t^{\mathrm{pr}}
+
\sum_{j=1}^{n}
P_j^{\mathsf T}
\Lambda_{j,t}
s_{j,t},
\label{eq:gauss_composed_system}
\end{equation}
with
\begin{equation}
\Lambda_{\mathrm{comp},t}
=
(1-n)
P_{\mathrm{sh}}^{\mathsf T}
\Lambda_t^{\mathrm{pr}}
P_{\mathrm{sh}}
+
\sum_{j=1}^{n}
P_j^{\mathsf T}
\Lambda_{j,t}
P_j.
\label{eq:gauss_composed_precision}
\end{equation}
The factor \(1-n\) implements the shared-prior correction from
Eq.~\eqref{eq:joint_posterior_shared_local}. No corresponding correction is
applied to the local blocks. The isotropic path-noise precision is counted
exactly once because
\begin{equation}
(1-n)
P_{\mathrm{sh}}^{\mathsf T}P_{\mathrm{sh}}
+
\sum_{j=1}^{n}
P_j^{\mathsf T}P_j
=
I.
\end{equation}
If the shared prior and all single-observation posteriors are Gaussian, and
their scores are evaluated exactly, this construction reproduces the exact
score of the perturbed joint posterior. In the absence of local parameter
blocks, it reduces to the standard tall-data \textsc{Gauss} composition of
Ref.~\cite{linhart2026diffusionposteriorsamplingsimulationbased}.

Because the shared-prior term enters with the negative coefficient \(1-n\),
\(\Lambda_{\mathrm{comp},t}\) is not automatically positive definite for an
arbitrary collection of approximate factors. We therefore monitor its
smallest eigenvalue during sampling and, when required, apply the minimum
diagonal shift needed to enforce a small positive eigenvalue floor.

Solving Eq.~\eqref{eq:gauss_composed_system} blockwise clarifies the roles of
the individual terms. Partition the denoising precision and score of
observation \(j\) as
\begin{equation}
\Lambda_{j,t}
=
\begin{pmatrix}
\Lambda^{\mathrm{ss}}_{j,t}
&
\Lambda^{\mathrm{sl}}_{j,t}
\\
\Lambda^{\mathrm{ls}}_{j,t}
&
\Lambda^{\mathrm{ll}}_{j,t}
\end{pmatrix},
\qquad
s_{j,t}
=
\begin{pmatrix}
s^{\mathrm{sh}}_{j,t}
\\
s^{\mathrm{loc}}_{j,t}
\end{pmatrix}.
\label{eq:block_precision_and_score}
\end{equation}
The local rows give
\begin{equation}
s^{\mathrm{loc}}_{\mathrm{comp},j,t}
=
s^{\mathrm{loc}}_{j,t}
+
\left(
\Lambda^{\mathrm{ll}}_{j,t}
\right)^{-1}
\Lambda^{\mathrm{ls}}_{j,t}
\left(
s^{\mathrm{sh}}_{j,t}
-
s^{\mathrm{sh}}_{\mathrm{comp},t}
\right).
\label{eq:composed_local_score}
\end{equation}
Thus, when composition moves the shared parameters away from the values
preferred by observation \(j\) alone, the local parameters of that
observation are corrected along their posterior correlation directions.

Eliminating the local blocks yields the shared-block solution
\begin{equation}
s^{\mathrm{sh}}_{\mathrm{comp},t}
=
\left[
(1-n)\Lambda_t^{\mathrm{pr}}
+
\sum_{j=1}^{n}
\bar\Lambda_{j,t}
\right]^{-1}
\left[
(1-n)\Lambda_t^{\mathrm{pr}}s_t^{\mathrm{pr}}
+
\sum_{j=1}^{n}
\bar\Lambda_{j,t}s^{\mathrm{sh}}_{j,t}
\right],
\label{eq:composed_shared_score}
\end{equation}
where
\begin{equation}
\bar\Lambda_{j,t}
=
\Lambda^{\mathrm{ss}}_{j,t}
-
\Lambda^{\mathrm{sl}}_{j,t}
\left(
\Lambda^{\mathrm{ll}}_{j,t}
\right)^{-1}
\Lambda^{\mathrm{ls}}_{j,t}.
\label{eq:shared_schur_precision}
\end{equation}
The Schur complement \(\bar\Lambda_{j,t}\) is the marginal precision with
which observation \(j\) constrains the shared parameters after accounting for
uncertainty in its local parameters. Each measurement therefore contributes
to the shared block in proportion to the information it actually carries
about that block, while the shared prior is counted exactly once.

For two observations, Eq.~\eqref{eq:composed_shared_score} reduces to
\begin{equation}
s^{\mathrm{sh}}_{\mathrm{comp},t}
=
\left(
\bar\Lambda_{1,t}
+
\bar\Lambda_{2,t}
-
\Lambda_t^{\mathrm{pr}}
\right)^{-1}
\left(
\bar\Lambda_{1,t}s^{\mathrm{sh}}_{1,t}
+
\bar\Lambda_{2,t}s^{\mathrm{sh}}_{2,t}
-
\Lambda_t^{\mathrm{pr}}s_t^{\mathrm{pr}}
\right),
\label{eq:composed_shared_score_two}
\end{equation}
which is the form presented in the Results.

If all shared--local off-diagonal entries of the factor precisions are
discarded, Eqs.~\eqref{eq:composed_local_score} and
\eqref{eq:composed_shared_score} reduce to an element-wise
precision-weighted average on the shared block and to
\begin{equation}
s^{\mathrm{loc}}_{\mathrm{comp},j,t}
=
s^{\mathrm{loc}}_{j,t}
\end{equation}
on the local blocks. This diagonal-covariance simplification ignores the
posterior correlations between shared and local parameters and, in
particular, sets the local correction in
Eq.~\eqref{eq:composed_local_score} to zero. We therefore use the
full-covariance rule throughout.

Once the composed score has been formed, it is converted back into a global
flow matching vector field using the inverse of
Eq.~\eqref{eq:general_score_from_flow},
\begin{equation}
v_{\mathrm{comp}}\!\left(
\Theta_t,t;C_{1:n}
\right)
=
\frac{
\dot m_t\Theta_t
+
\sigma_t
\left(
\dot m_t\sigma_t-m_t\dot\sigma_t
\right)
s_{\mathrm{comp},t}
}{
m_t
}.
\label{eq:global_vector_field_from_score_general}
\end{equation}
For the coefficients used here, this becomes
\begin{equation}
v_{\mathrm{comp}}\!\left(
\Theta_t,t;C_{1:n}
\right)
=
\frac{
\Theta_t
+
(1-t+\sigma_{\min})(1+\sigma_{\min})
s_{\mathrm{comp},t}
}{
t
}.
\label{eq:global_vector_field_from_score}
\end{equation}
Posterior samples are generated by integrating the global ordinary
differential equation
\begin{equation}
\frac{\mathrm d\Theta_t}{\mathrm dt}
=
v_{\mathrm{comp}}\!\left(
\Theta_t,t;C_{1:n}
\right).
\label{eq:global_compositional_ode}
\end{equation}
starting from a small non-zero time \(t_{\min}=0.001\) to avoid the singularity
in Eq.~\eqref{eq:global_vector_field_from_score}.

This formulation covers two practically important cases. In the homogeneous case, all observations use the same parameterization and the same trained conditional flow model. This setting is natural for multi-contrast neutron reflectometry, where structural parameters are shared across contrasts while contrast-dependent or instrumental parameters remain observation-specific. In the heterogeneous case, different observations may use different local parameterizations and different trained conditional flow models,  provided that the shared parameters represent the same physical quantities and use identical inference-time prior bounds and probability paths.

\subsection*{MCMC refinement}

Raw amortized posteriors produced by neural network models in SBI approaches are generally approximate. We therefore optionally apply a second-stage posterior refinement step after amortized sampling. The purpose of this step is to target the posterior defined by the explicit forward-model likelihood and the selected prior bounds using the amortized posterior as an initialization distribution.

Specifically, we use samples generated by the neural network model to initialize an ensemble Markov chain Monte Carlo (MCMC) sampler. For refinement, we use a GPU-parallel affine-invariant ensemble MCMC scheme~\cite{goodman2010ensemble}, adapted from the PANPE framework~\cite{vladimir_panpe}. 

This refinement step combines the speed of amortized neural inference with likelihood-based posterior sampling. In contrast to classical initialization strategies based on a single best-fit solution with a small local perturbation, amortized posterior samples provide an ensemble of starting points already located in plausible regions of parameter space. This is particularly useful when the posterior is non-Gaussian, correlated or multimodal.

\subsection*{Training data generation}

Training data are generated on the fly using a batched GPU implementation of the reflectometry forward model. At each training iteration, physical parameter values and corresponding prior bounds are sampled from the model-specific training ranges (listed in the Supplementary Information). The sampling procedure follows previous work on prior-conditioned point estimation and posterior estimation models~\cite{Munteanu2024,vladimir_panpe}. For slab-style parameterizations, we enforce physically motivated constraints, such as interfacial roughnesses not exceeding half the thicknesses of adjacent layers and the total layer thickness not exceeding \(1500\,\text{\AA}\). For the XRR parameterization with absorption, the imaginary part of the electronic SLD is restricted not to exceed 0.2 times the corresponding real part to avoid strongly unphysical training examples.

The \(q\) points used to simulate reflectivity curves are sampled from predefined ranges for the minimum and maximum \(q\) value and for the number of points. For XRR models, the \(q\) spacing is uniform. For NR models, half of each training batch uses uniformly spaced points and half uses randomly spaced points. This choice makes the NR models more robust to non-uniformly sampled data, as commonly encountered in time-of-flight measurements.

Theoretical XRR or NR curves are simulated from the sampled \(q\) values and structural parameters using the Abel\`es transfer-matrix formalism~\cite{Abeles1950}, as implemented in the \texttt{reflectorch} package~\cite{Munteanu2025}. Non-slab parameterizations, such as the polymer-brush model, are first converted to an equivalent slab representation. For non--spin-flip PNR, the two reflectivity channels are simulated separately from the effective SLDs
\(\rho^{\uparrow}=\rho^n+\rho^m\) and
\(\rho^{\downarrow}=\rho^n-\rho^m\), where \(\rho^n\) and \(\rho^m\) denote the nuclear and magnetic SLDs.

For full spin-resolved PNR, we use a transfer-matrix method with \(4\times4\) matrices~\cite{Blundell1992}, implemented in PyTorch for this work and verified against \texttt{refnx}. In this setting, interfacial roughness cannot be incorporated through the usual N\'evot--Croce factor used with the Abel\`es formalism. A more detailed treatment would require microslicing, which is computationally expensive for large-scale neural-network training. We therefore restrict the full spin-resolved PNR example to a simplified parameterization without roughness. Additional PNR effects, such as magnetic-domain roughness, Zeeman effects or helical magnetic structures, are outside the scope of the present work~\cite{Singh2009,Maranville2016,Karhu2011}.

For NR models, instrumental resolution smearing is included during training. We apply linear resolution smearing with coefficient \(dq/q\), sampled from a predefined range for each curve and supplied to the neural network as an additional conditioning variable. This provides an approximation to pointwise resolution smearing in time-of-flight data; when pointwise resolution values are available experimentally, they are used explicitly during MCMC refinement. Because resolution smearing requires simulation on a finer \(q\)-grid followed by convolution and interpolation, NR simulations for training are more expensive than XRR simulations in both computation time and memory usage.

Finally, the simulated curves are modified by sampled instrumental parameters, which may include a \(q\)-shift, intensity scale factor or constant background. For non--spin-flip PNR, the theoretical channels may also be mixed by a coefficient \(\alpha\), which provides a phenomenological channel-leakage correction for imperfect neutron-beam polarization. More complete treatments of polarization analysis describe imperfect polarizers, flippers and analyzers through efficiency matrices~\cite{Wildes1999}. Gaussian noise is then added to the simulated curves.

\subsection*{Model training and inference settings}

All neural networks were trained using the prior-amortized flow matching objective with a normal base distribution, two-sided conditioning and linear probability paths. We used the AdamW optimizer for \(500\,000\) training iterations, which took approximately one day on a single NVIDIA A100 GPU for the model sizes used here. The learning rate followed a cosine annealing schedule from \(1.0\times10^{-3}\), after a 500-step warm-up, to \(1.0\times10^{-6}\). Unless stated otherwise, posterior samples from trained FMPE models were generated by integrating the learned ODE with the adaptive Dormand--Prince solver implemented in \texttt{torchdiffeq}. Training was performed on NVIDIA A100 GPUs with dedicated FP64 cores for double-precision arithmetic. Because the reflectometry simulations rely on double-precision (\texttt{float64}) arithmetic, their efficiency is significantly improved on GPUs that provide high throughput for double-precision operations.

\subsection*{Polymer-brush parameterization}

For the polymer-brush example, we use a structured parameterization for an end-grafted brush following classical brush theory~\cite{Milner1988}. The polymer volume-fraction profile is
\begin{equation}
\phi(z) = \phi_0 \left[1 - \left(\frac{z}{L}\right)^n\right],
\qquad 0 \le z \le L,
\end{equation}
where \(\phi_0\) is the polymer volume fraction at the grafting interface, \(L\) is the brush height and \(n\) is the brush-profile exponent. The case \(n=2\) corresponds to an ideal parabolic brush profile, while other values allow deviations from that ideal form.

To evaluate the reflectivity, the continuous brush profile is discretized into \(N=10\) slices of equal thickness,
\begin{equation}
\Delta z = \frac{L}{N}.
\end{equation}
Each slice is assigned an effective SLD through linear mixing of polymer and solvent,
\begin{equation}
\rho_i = \phi_i \rho_{\mathrm{pol}} + \left(1-\phi_i\right)\rho_{\mathrm{solv}},
\end{equation}
where \(1-\phi_i\) is the solvent volume fraction. The roughness between adjacent brush slices is fixed to
\begin{equation}
\sigma_i = \frac{L}{2N},
\end{equation}
which smooths the discretized representation of the continuous profile, following the implementation used in \texttt{anaklasis}~\cite{Koutsioubas2021}. A thin reactive layer of thickness \(d_{\mathrm{rl}}\) and SLD \(\rho_{\mathrm{rl}}\) is included between the fronting medium and the brush.

\section*{Data availability}

The experimental datasets analyzed in this study are available from the public sources cited in the corresponding Results sections. The time-of-flight NR dataset is publicly available from the Institut Laue--Langevin data archive at
\url{https://doi.ill.fr/10.5291/ILL-DATA.9-11-1903}. The polymer-brush and XRR/PNR co-refinement datasets are distributed along with the \texttt{anaklasis} package~\cite{Koutsioubas2021}. The multiple-contrast NR dataset is distributed along with the \texttt{refnx} examples~\cite{NelsonRefnx}. The synthetic full spin-resolved PNR dataset was generated using the forward model and parameter settings described in Methods. 

\section*{Code availability}

The code used to train the flow matching posterior estimators, perform posterior sampling, run MCMC refinement and generate the figures will be made available as part of the \texttt{reflectorch} package. A version of the code associated with this manuscript, together with configuration files and scripts required to reproduce the reported analyses, will be archived in a DOI-minting public repository before publication.

\section*{Author contributions}
V.M. implemented the machine learning methods introduced in this work, trained the neural networks, evaluated the results and wrote the first draft of the paper. V.S. provided expertise on the machine learning methods. D.L., A.H. and F.S. provided expertise on reflectometry experiments and supervised the project. All authors contributed to refining the text of the paper.

\section*{Competing interests}

The authors declare no competing interests.

\section*{Acknowledgements}

This work was supported by the DFG (project number 557510229). This research is part of the VIPR project (05D23VT1, ERUM-DATA), funded by the German Federal Ministry for Education and Research (BMBF). We acknowledge support by the consortium DAPHNE4NFDI in the context of the work of NFDI e.V., funded by the German Research Foundation (DFG, grant number 460248799). We thank the University of Tübingen Cluster of Excellence ``Machine Learning: New Perspectives for Science'' for support (EXC 2064/1, project no. 390727645). We acknowledge DESY (Hamburg, Germany), a member of the Helmholtz Association, for the provision of experimental facilities. We thank Dr. Alexandros Koutsioubas for making the experimental data from the \texttt{anaklasis} package available.

\bibliography{references_reflectivity}

\clearpage

\renewcommand{\thetable}{\arabic{table}}
\begin{center}
  {\Large\bfseries Supplementary Information\par}
\end{center}
\vspace{1em}

\begin{figure}[h!]
  \centering
  \includegraphics[width=1.0\linewidth]{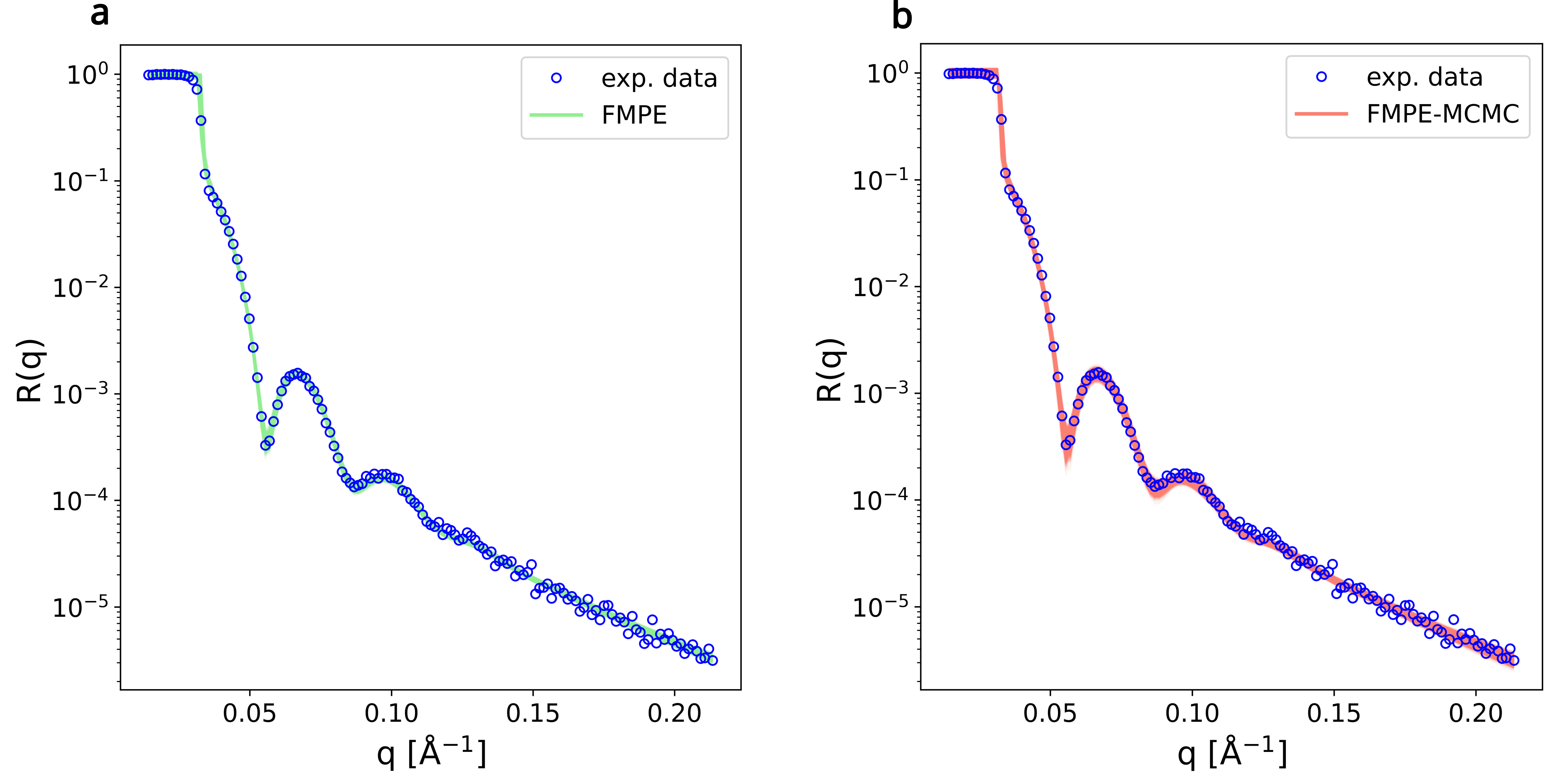}
  \caption{Experimental XRR data for a one-layer PTCDI-C$_{3}$ film on a Si substrate are shown together with posterior-predictive simulations generated from parameter samples.
  (a) Posterior-predictive curves obtained directly from the raw FMPE posterior.
  (b) Posterior-predictive curves obtained after FMPE-initialized MCMC refinement.}
  \label{fig:SI_XRR_PTCDI-C3}
\end{figure}

\begin{figure}[tbp]
  \centering
  \includegraphics[width=\linewidth]{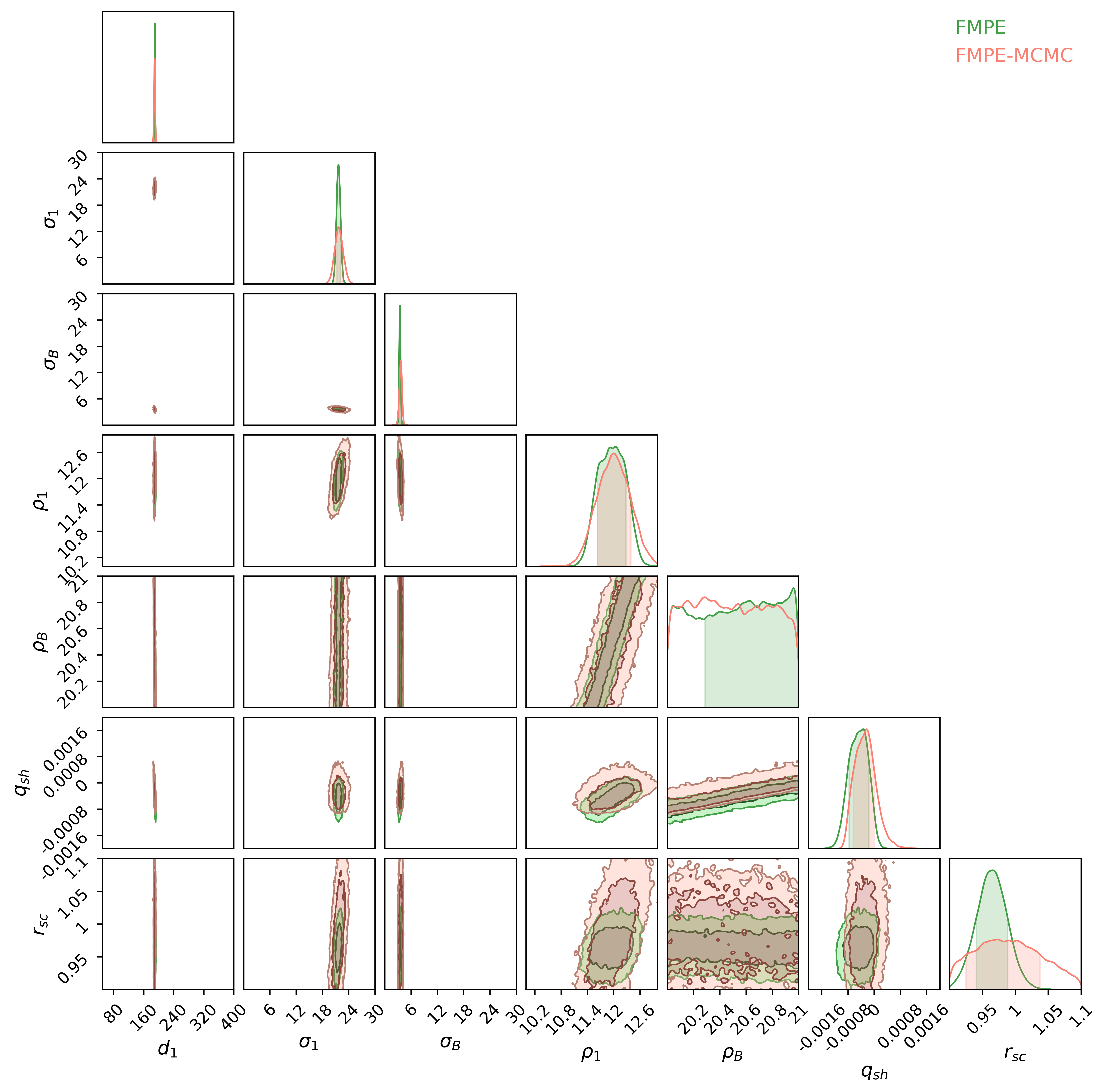}
  \caption{Corner plot of the raw FMPE (green) and FMPE-MCMC (red) posterior inferred for the XRR data of the PTCDI-C$_{3}$ film. Filled contours show the 68\% and 95\% credible regions. The displayed parameters comprise layer thicknesses \(d_i\), interfacial roughnesses \(\sigma_i\), SLDs \(\rho_i\), the \(q\)-shift offset \(q_{\mathrm{sh}}\) and the reflectivity scale factor \(r_{\mathrm{sc}}\). Units: \(d_i\) and \(\sigma_i\) are in \(\mathrm{\AA}\), \(\rho_i\) in \(10^{-6}\,\mathrm{\AA}^{-2}\), \(q_{\mathrm{sh}}\) in \(\mathrm{\AA}^{-1}\), and \(r_{\mathrm{sc}}\) is dimensionless. The subscript B denotes the backing medium. The plot ranges correspond to the prior bounds.}
  \label{fig:XRR_PTCDI-C3_cornerplot_mcmc}
\end{figure}

\begin{figure}[h!]
  \centering
  \includegraphics[width=1.0\linewidth]{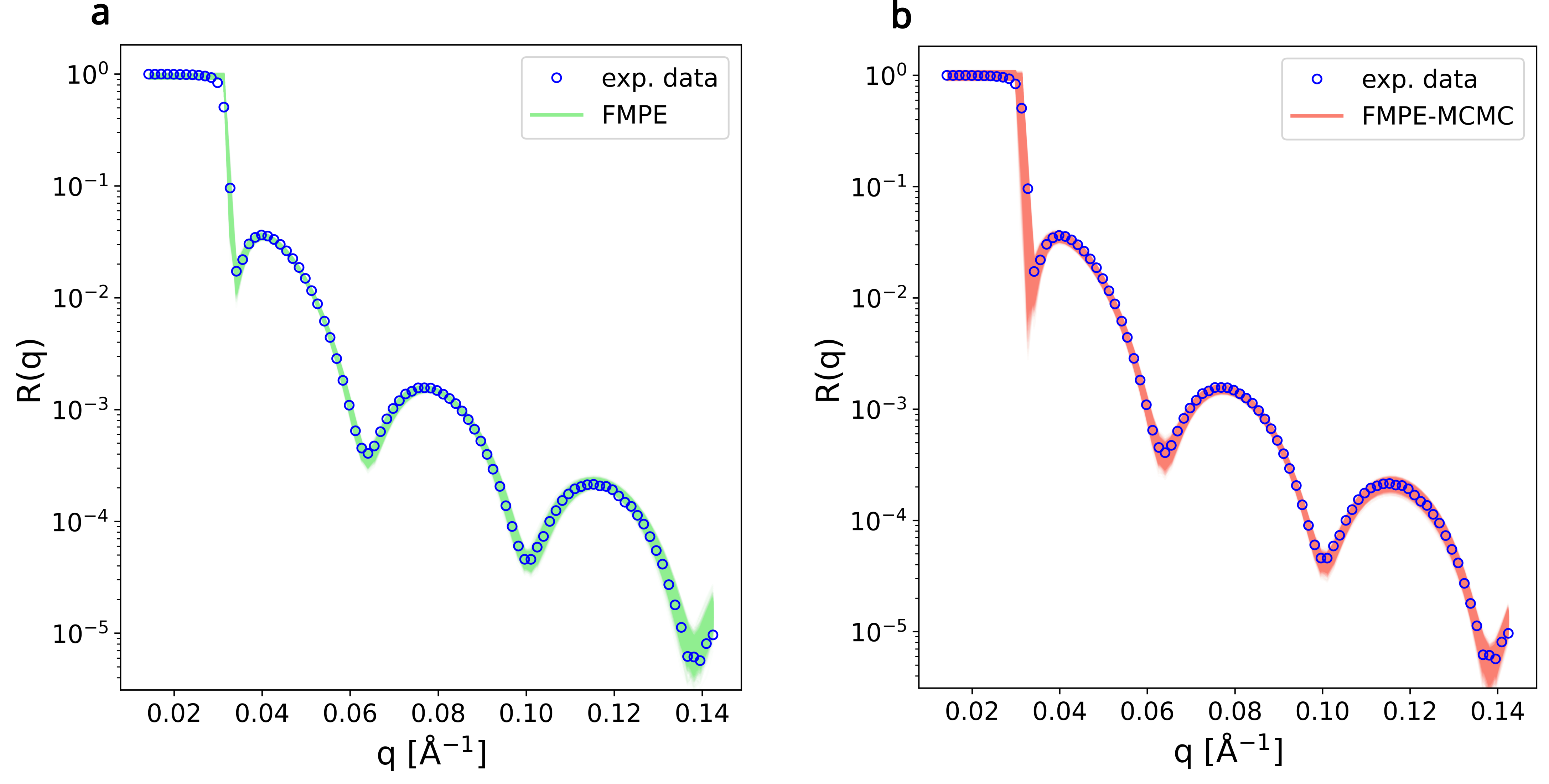}
  \caption{Experimental XRR data for a one-layer C$_{60}$ film on a Si substrate are shown together with posterior-predictive simulations generated from parameter samples.
  (a) Posterior-predictive curves obtained directly from the raw FMPE posterior.
  (b) Posterior-predictive curves obtained after FMPE-initialized MCMC refinement.}
  \label{fig:SI_XRR_C60}
\end{figure}

\begin{figure}[tbp]
  \centering
  \includegraphics[width=\linewidth]{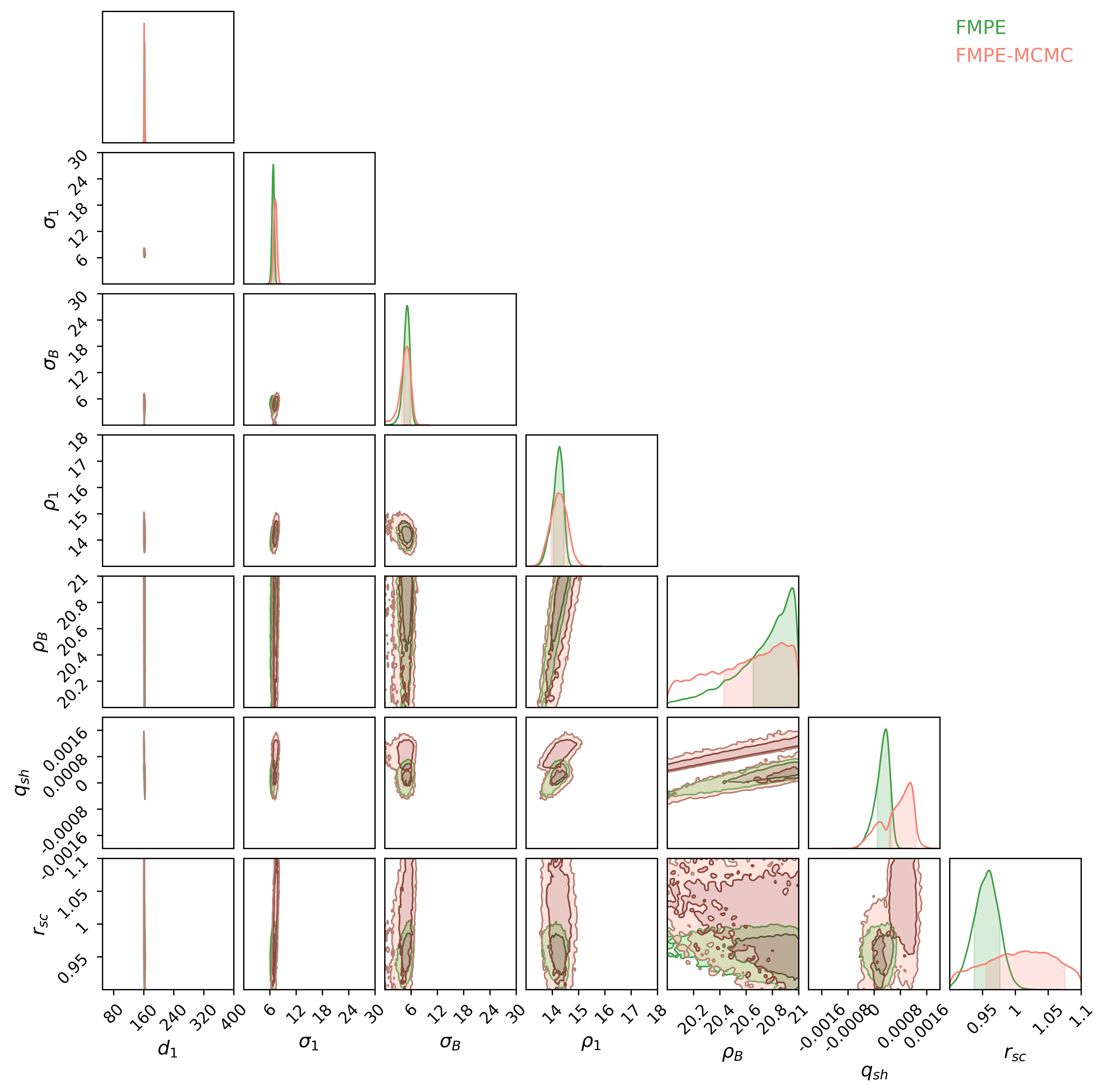}
  \caption{Corner plot of the raw FMPE (green) and FMPE-MCMC (red) posterior inferred for the XRR data of the C$_{60}$ film. Filled contours show the 68\% and 95\% credible regions. The displayed parameters comprise layer thicknesses \(d_i\), interfacial roughnesses \(\sigma_i\), SLDs \(\rho_i\), the \(q\)-shift offset \(q_{\mathrm{sh}}\) and the reflectivity scale factor \(r_{\mathrm{sc}}\). Units: \(d_i\) and \(\sigma_i\) are in \(\mathrm{\AA}\), \(\rho_i\) in \(10^{-6}\,\mathrm{\AA}^{-2}\), \(q_{\mathrm{sh}}\) in \(\mathrm{\AA}^{-1}\), and \(r_{\mathrm{sc}}\) is dimensionless. The subscript B denotes the backing medium. The plot ranges correspond to the prior bounds.}
  \label{fig:XRR_C60_cornerplot_mcmc}
\end{figure}

\begin{figure}[h!]
  \centering
  \includegraphics[width=1.0\linewidth]{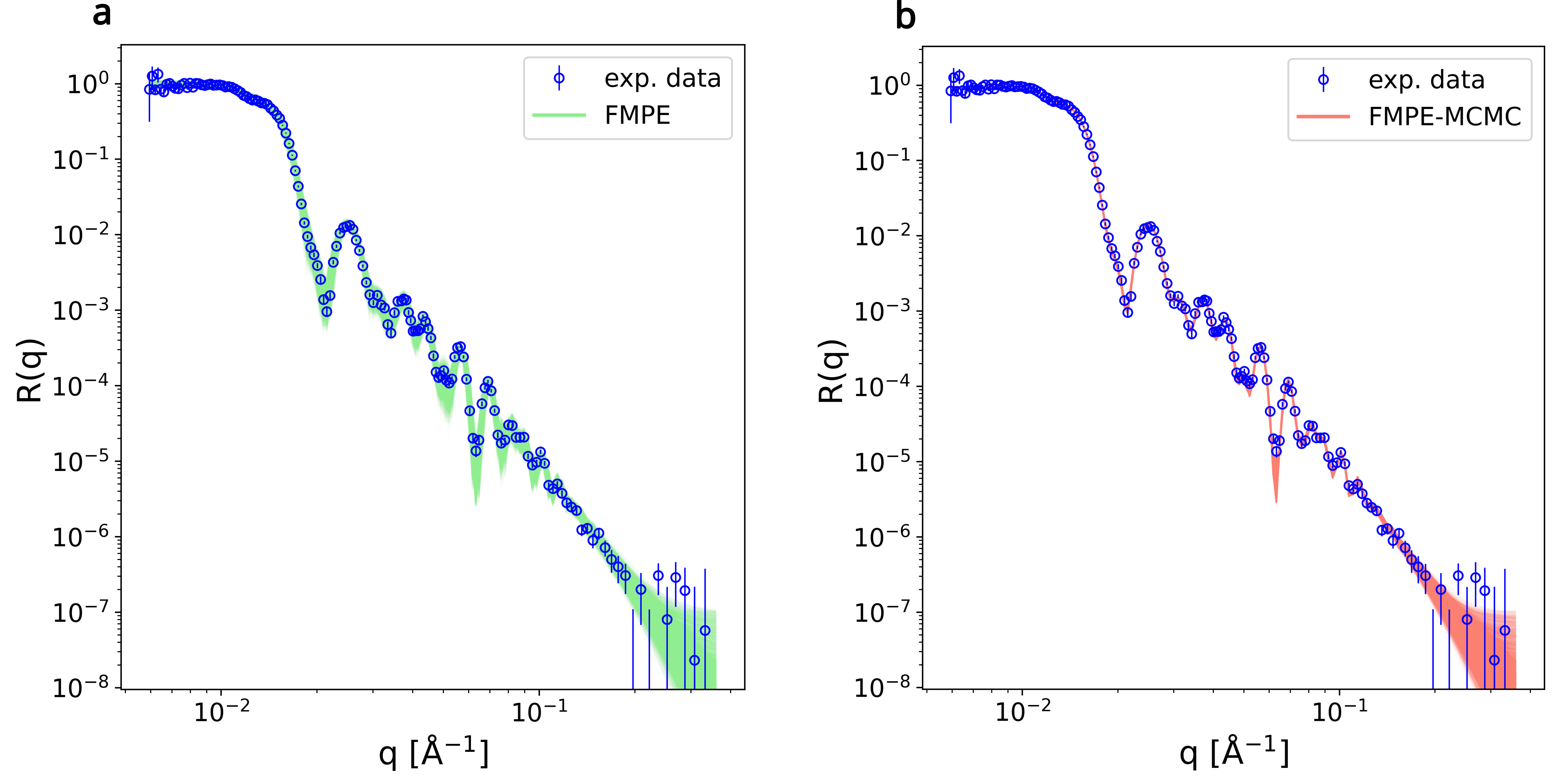}
  \caption{Experimental TOF-NR data for the PS:Bis-PCBM/Bis-PCBM structure, shown together with posterior-predictive simulations generated from parameter samples.
  (a) Posterior-predictive curves obtained directly from the raw FMPE posterior.
  (b) Posterior-predictive curves obtained after FMPE-initialized MCMC refinement.}
  \label{fig:SI_NR_ILL}
\end{figure}

\begin{figure}[tbp]
  \centering
  \includegraphics[width=\linewidth]{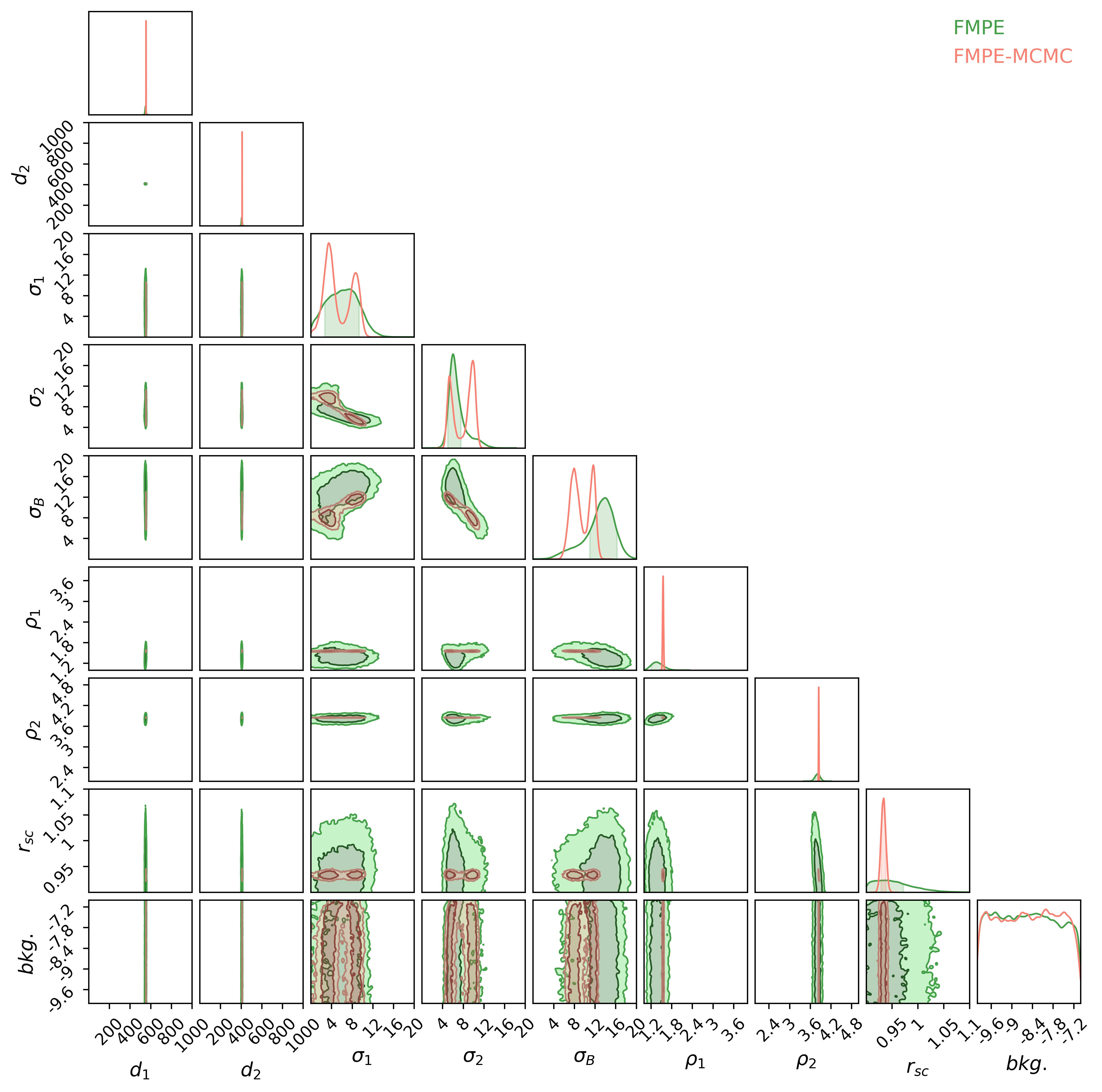}
  \caption{Corner plot of the raw FMPE (green) and FMPE-MCMC (red) posterior inferred for the NR data from the time-of-flight reflectometer. Filled contours show the 68\% and 95\% credible regions. The displayed parameters comprise layer thicknesses \(d_i\), interfacial roughnesses \(\sigma_i\), SLDs \(\rho_i\), the reflectivity scale factor \(r_{\mathrm{sc}}\) and the \(\log_{10}\) background. Units: \(d_i\) and \(\sigma_i\) are in \(\mathrm{\AA}\), \(\rho_i\) in \(10^{-6}\,\mathrm{\AA}^{-2}\), while \(r_{\mathrm{sc}}\) and the \(\log_{10}\) background are dimensionless. The subscript B denotes the backing medium. The plot ranges correspond to the prior bounds.}
\label{fig:NR_ILL_479365_479366_cornerplot_mcmc}
\end{figure}

\begin{figure}[h!]
  \centering
  \includegraphics[width=1.0\linewidth]{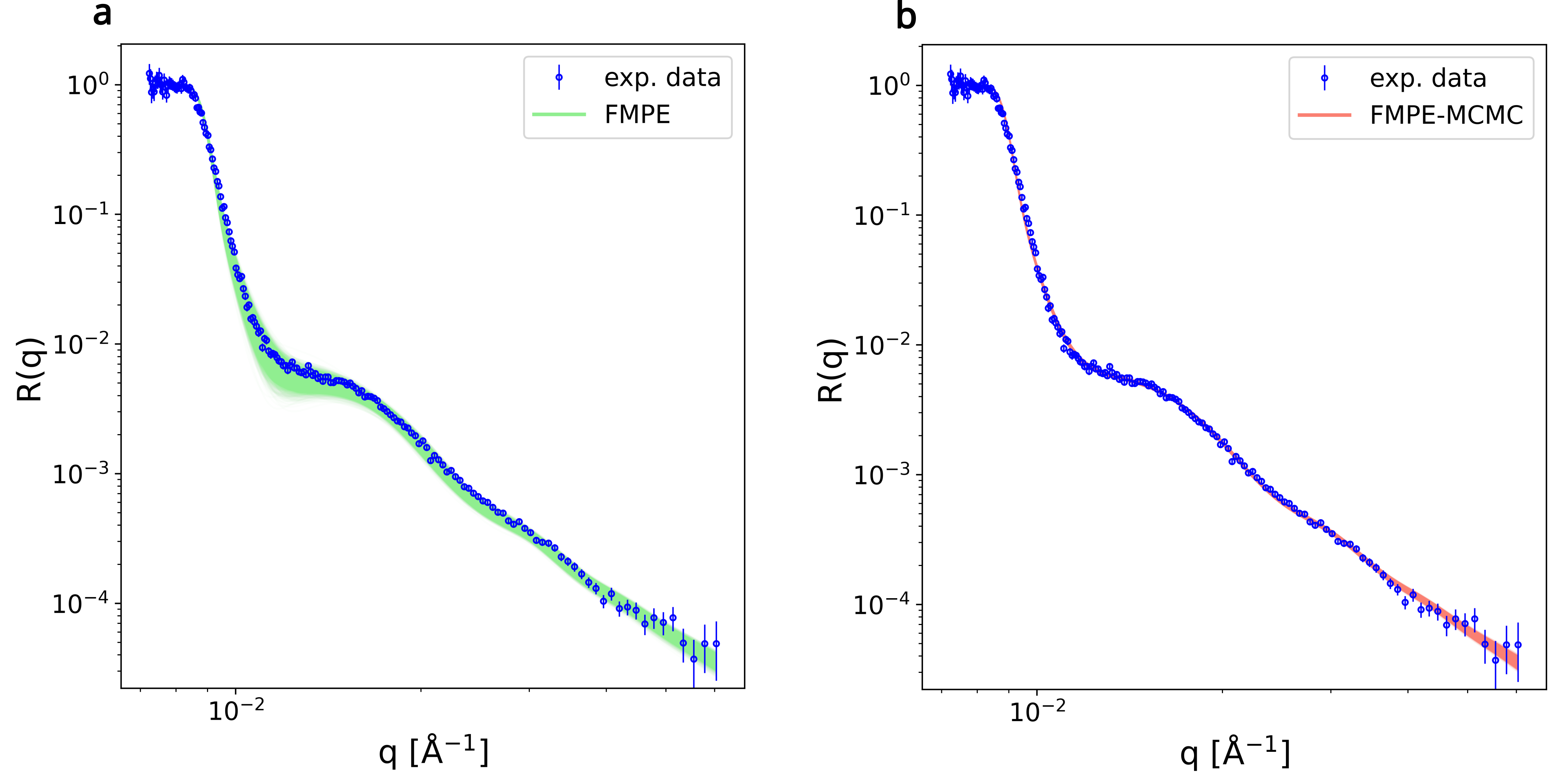}
  \caption{Experimental NR data for the polymer-brush structure, shown together with posterior-predictive simulations generated from parameter samples.
  (a) Posterior-predictive curves obtained directly from the raw FMPE posterior.
  (b) Posterior-predictive curves obtained after FMPE-initialized MCMC refinement.}
  \label{fig:SI_NR_polymerbrush}
\end{figure}

\begin{figure}[tbp]
  \centering
  \includegraphics[width=\linewidth]{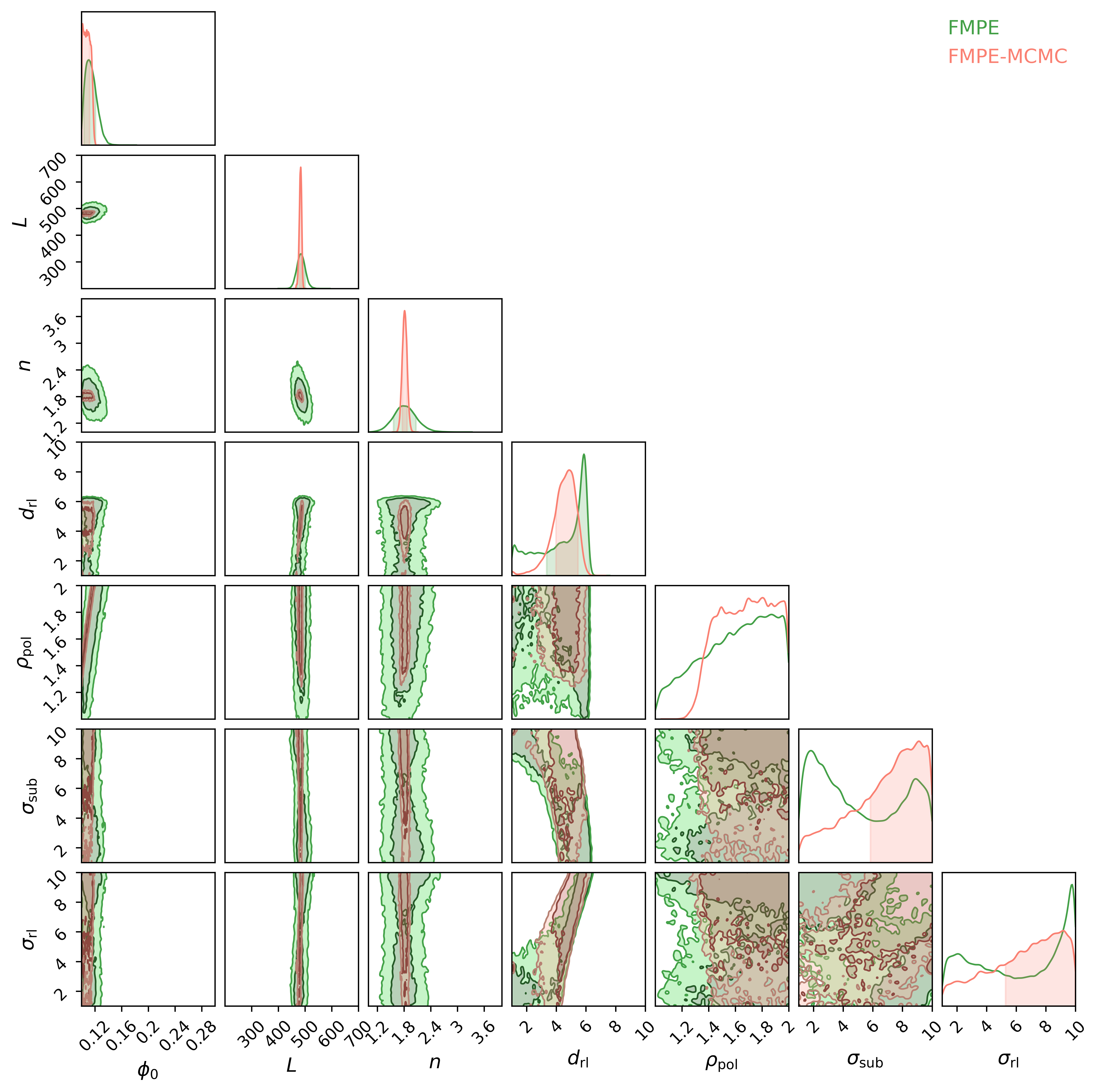}
  \caption{Corner plot of the raw FMPE (green) and FMPE-MCMC (red) posterior inferred for the NR data of the polymer-brush structure (data sourced from \cite{Koutsioubas2021}). Filled contours show the 68\% and 95\% credible regions. The displayed parameters comprise the polymer volume fraction at the grafting interface \(\phi_0\), the length of the brush \(L\), the polymer-brush exponent \(n\), the thickness of the reactive layer \(d_{\mathrm{rl}}\), the SLD of the polymer \(\rho_{\mathrm{pol}}\), the roughness of the substrate \(\sigma_{\mathrm{sub}}\) and the roughness of the reactive layer \(\sigma_{\mathrm{rl}}\). Units: \(L\), \(d_{\mathrm{rl}}\), \(\sigma_{\mathrm{sub}}\) and \(\sigma_{\mathrm{rl}}\) are in \(\mathrm{\AA}\), \(\rho_{\mathrm{pol}}\) is in \(10^{-6}\,\mathrm{\AA}^{-2}\), and \(\phi_0\) and \(n\) are dimensionless. The plot ranges correspond to the prior bounds.}
  \label{fig:NR_polymerbrush_cornerplot_mcmc}
\end{figure}

\begin{figure}[h!]
  \centering
  \includegraphics[width=1.0\linewidth]{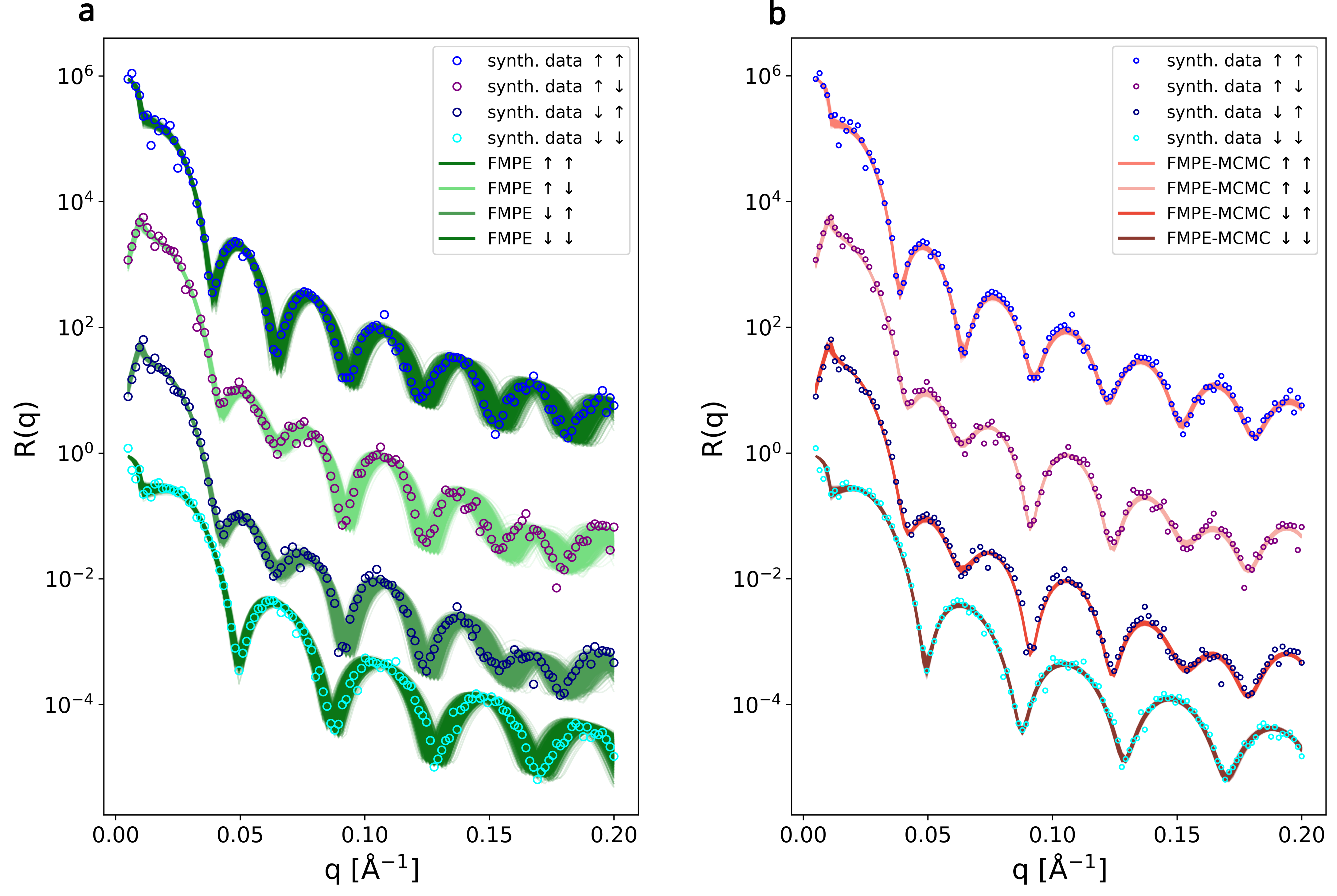}
  \caption{Synthetic full spin-resolved PNR data, shown together with posterior-predictive simulations generated from parameter samples.
  (a) Posterior-predictive curves obtained directly from the raw FMPE posterior.
  (b) Posterior-predictive curves obtained after FMPE-initialized MCMC refinement.}
  \label{fig:SI_PNR_spinflip}
\end{figure}

\begin{figure}[tbp]
  \centering
  \includegraphics[width=\linewidth]{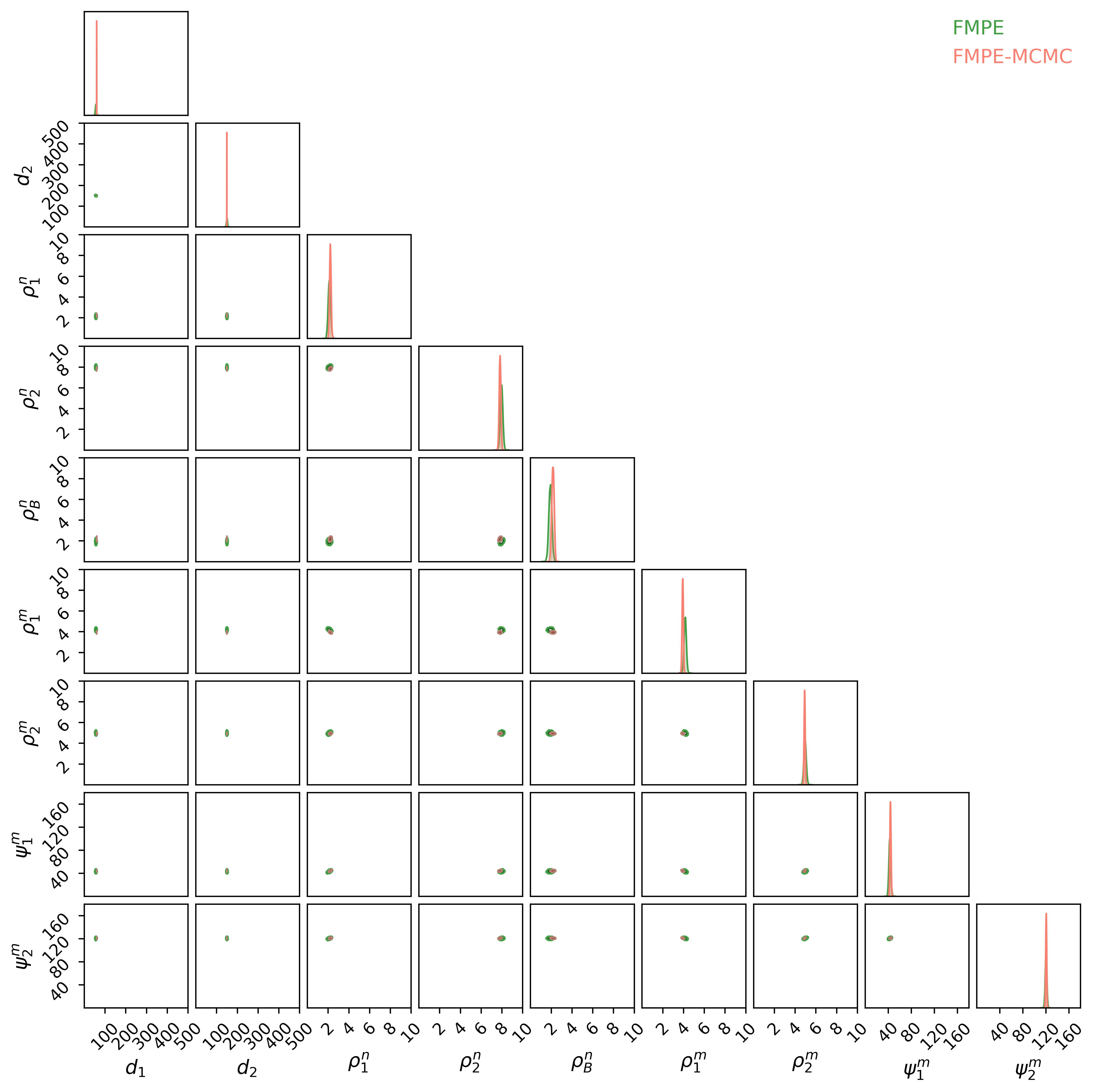}
  \caption{Corner plot of the raw FMPE (green) and FMPE-MCMC (red) posterior inferred for the synthetic full spin-resolved PNR data. The displayed parameters comprise layer thicknesses \(d_i\), nuclear SLDs \(\rho^{n}_{i}\), magnetic SLDs \(\rho^{m}_{i}\) and magnetization angles \(\psi^{m}_{i}\). Units: \(d_i\) are in \(\mathrm{\AA}\), \(\rho^{n}_{i}\) and \(\rho^{m}_{i}\) are in \(10^{-6}\,\mathrm{\AA}^{-2}\), and \(\psi^{m}_{i}\) is in degrees. The subscript B denotes the backing medium. The plot ranges correspond to the prior bounds.}
  \label{fig:PNR_spinflip_ex1_cornerplot}
\end{figure}

\begin{figure}[h!]
  \centering
  \includegraphics[width=1.0\linewidth]{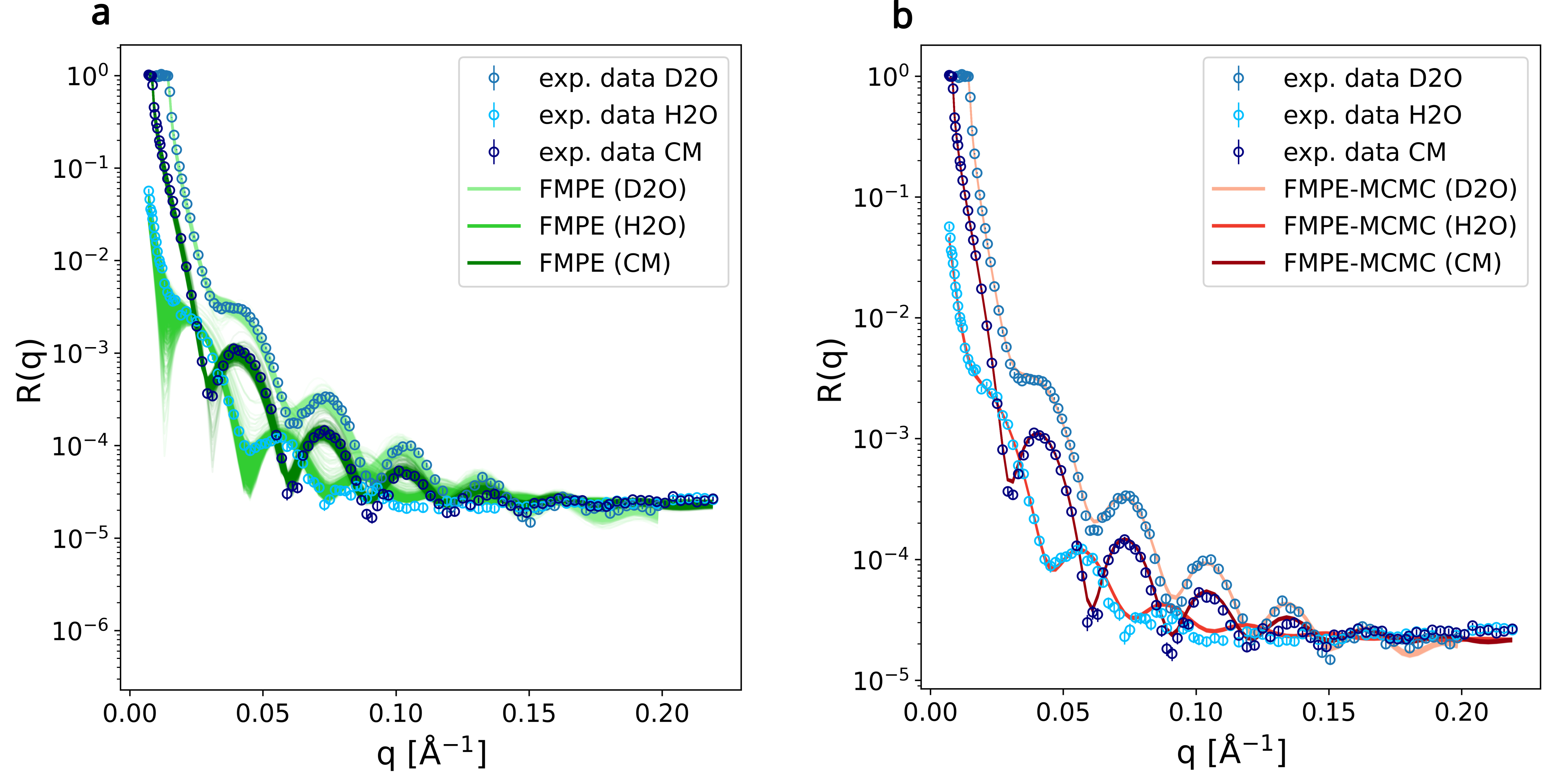}
  \caption{Experimental NR data corresponding to three solvent contrasts (D$_2$O, H$_2$O and CM), shown together with posterior-predictive simulations generated from parameter samples.
  (a) Posterior-predictive curves obtained directly from the raw FMPE posterior.
  (b) Posterior-predictive curves obtained after FMPE-initialized MCMC refinement.}
  \label{fig:SI_NR_slab_contrasts}
\end{figure}

\begin{figure}[tbp]
  \centering
  \includegraphics[width=\linewidth]{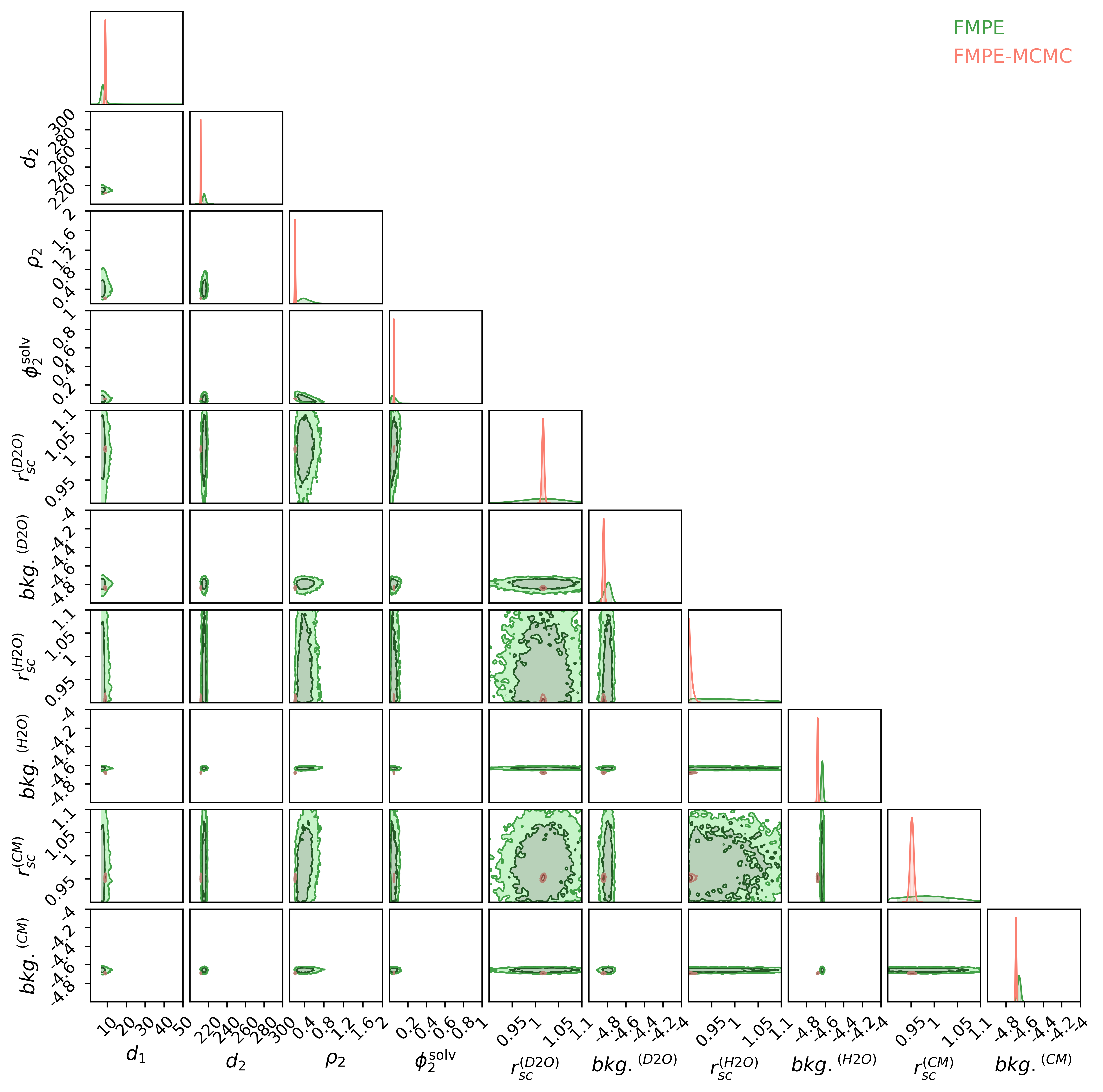}
  \caption{Corner plot of the raw FMPE (green) and FMPE-MCMC (red) posterior inferred for the three-contrast NR data. Filled contours show the 68\% and 95\% credible regions. The displayed parameters comprise the layer thicknesses \(d_i\), the SLD of the polymer \(\rho_2\) without the solvent, the solvent volume fraction in the polymer layer \(\phi^{\mathrm{solv}}_2\) and the reflectivity scale factors \(r_{\mathrm{sc}}\) and \(\log_{10}\) backgrounds for each curve. Units: \(d_i\) are in \(\mathrm{\AA}\), \(\rho_2\) is in \(10^{-6}\,\mathrm{\AA}^{-2}\), and \(\phi^{\mathrm{solv}}_2\), \(r_{\mathrm{sc}}\) and the \(\log_{10}\) backgrounds are dimensionless. The plot ranges correspond to the prior bounds.}
  \label{fig:NR_slab_contrasts_cornerplot}
\end{figure}

\begin{figure}[h!]
  \centering
  \includegraphics[width=1.0\linewidth]{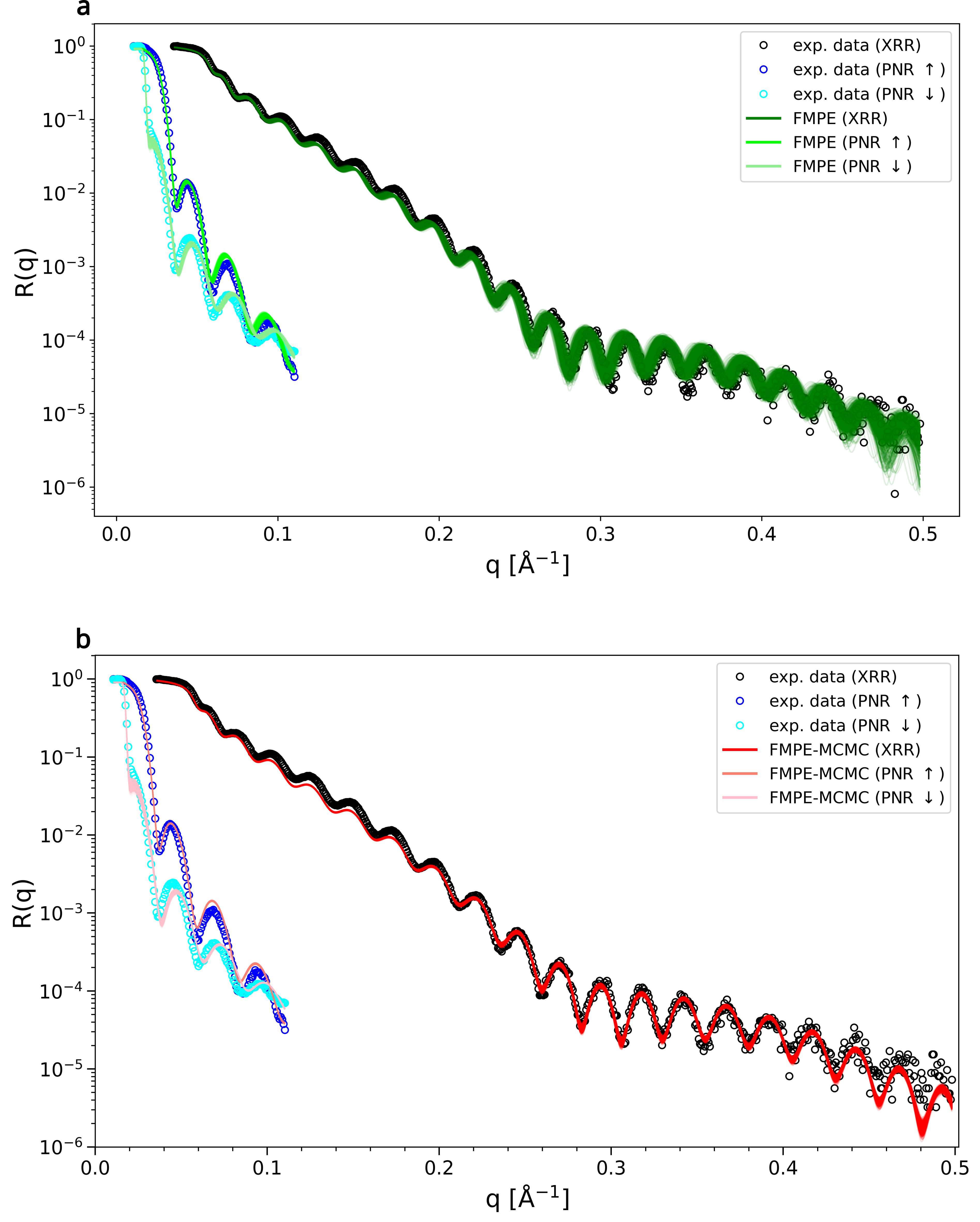}
  \caption{Experimental XRR and PNR data for the Pt/Fe structure, shown together with posterior-predictive simulations generated from parameter samples.
  (a) Posterior-predictive curves obtained directly from the raw FMPE posterior.
  (b) Posterior-predictive curves obtained after FMPE-initialized MCMC refinement.}
  \label{fig:SI_XRRandPNR_PtFe}
\end{figure}

\begin{figure}[tbp]
  \centering
  \includegraphics[width=\linewidth]{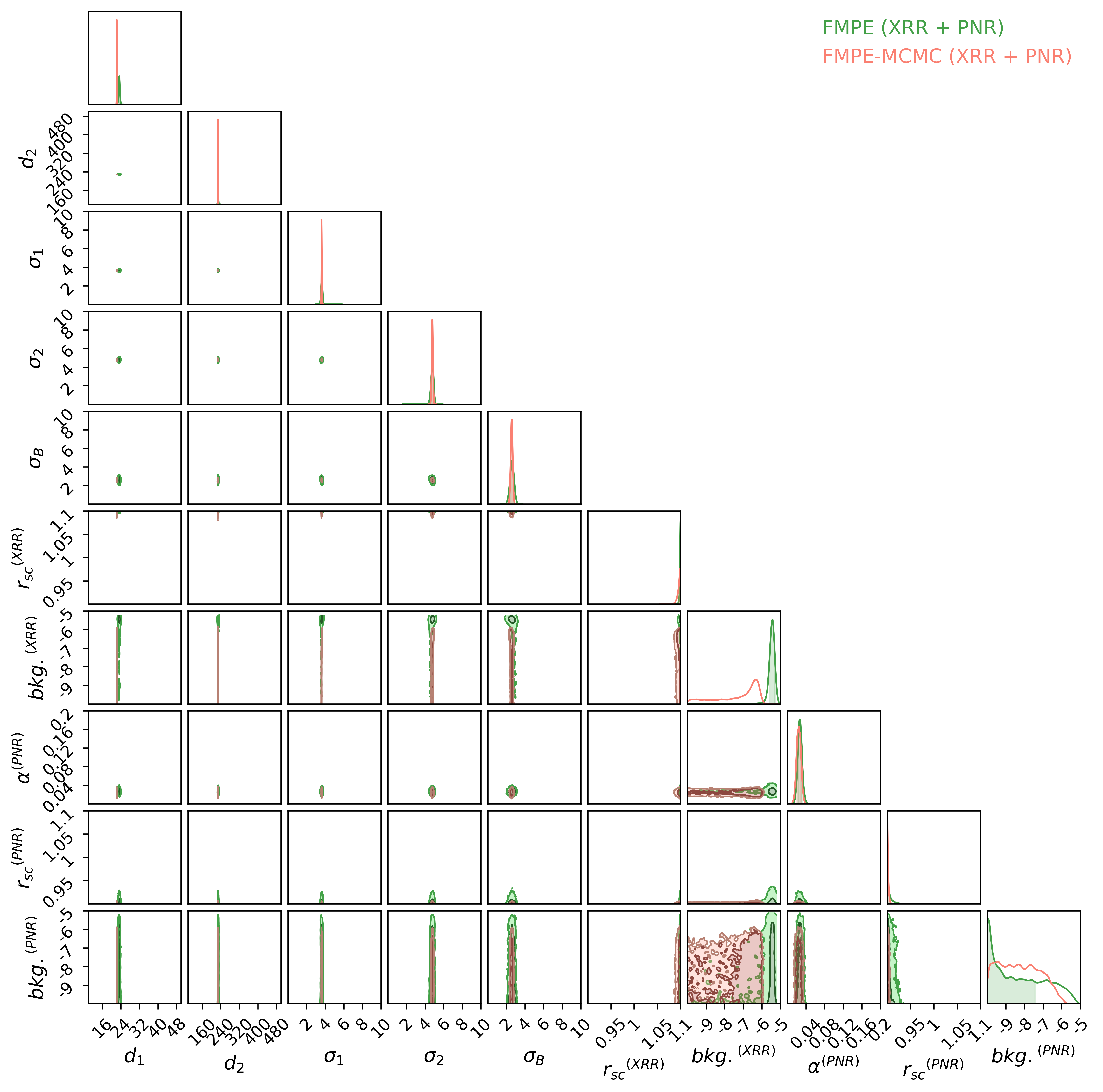}
  \caption{Corner plot of the raw FMPE (green) and FMPE-MCMC (red) posterior inferred for the Pt/Fe structure. Filled contours show the 68\% and 95\% credible regions. The displayed parameters comprise the layer thicknesses \(d_i\), interfacial roughnesses \(\sigma_i\), the reflectivity scale factors \(r_{\mathrm{sc}}\), the \(\log_{10}\) backgrounds and, for PNR, the mixing coefficient \(\alpha\). Units: \(d_i\) and \(\sigma_i\) are in \(\mathrm{\AA}\), while \(r_{\mathrm{sc}}\), the \(\log_{10}\) backgrounds and \(\alpha\) are dimensionless. The plot ranges correspond to the prior bounds.}
  \label{fig:PtFe_XRR_and_PNR_cornerplot}
\end{figure}

\begin{figure}[tbp]
  \centering
  \includegraphics[width=\linewidth]{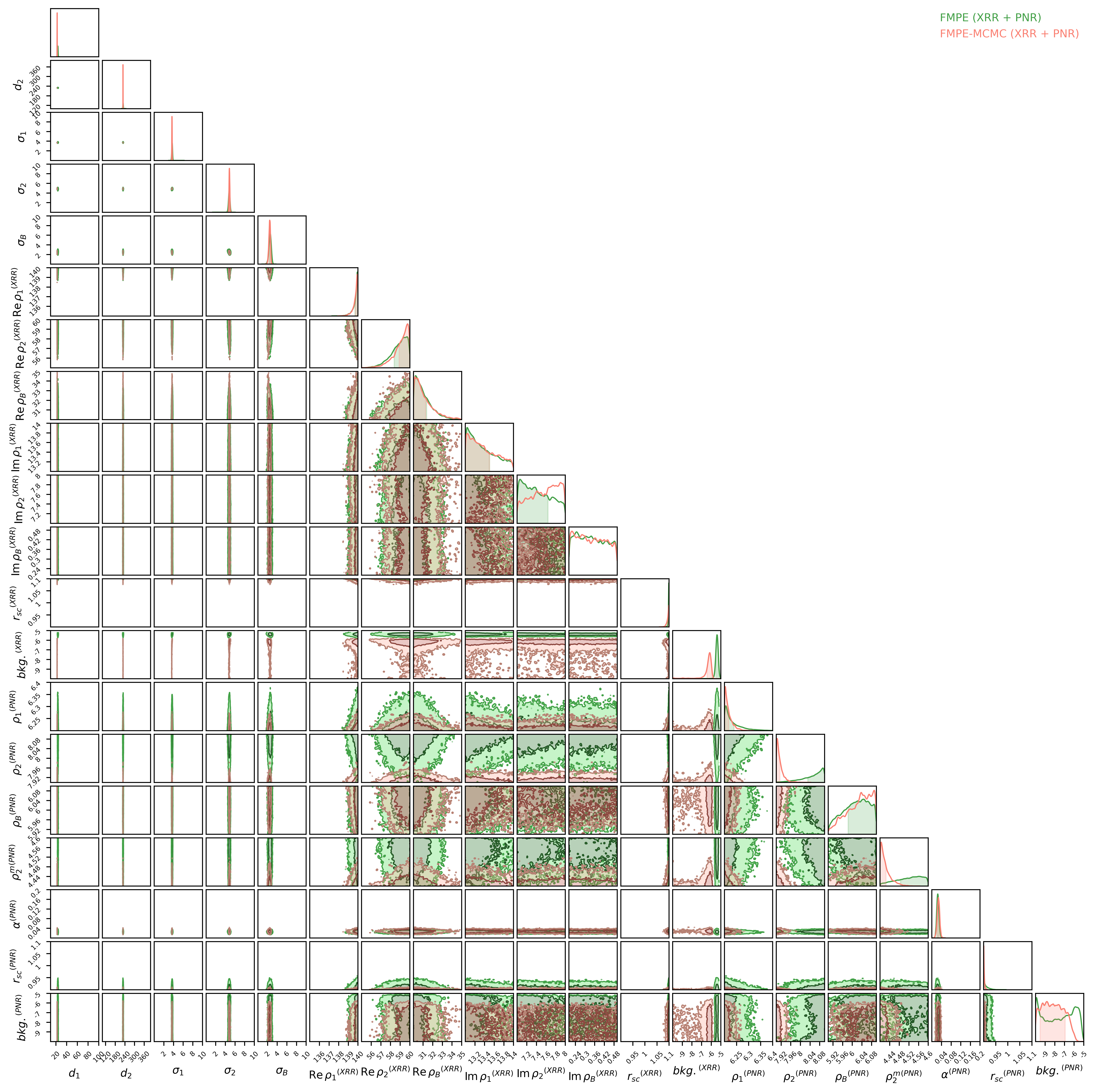}
  \caption{Corner plot of the raw FMPE (green) and FMPE-MCMC (red) posterior, in the case of an alternative inference where the SLDs are not fixed, for the Pt/Fe structure. Filled contours show the 68\% and 95\% credible regions. The displayed parameters comprise the layer thicknesses \(d_i\), interfacial roughnesses \(\sigma_i\), real electronic SLDs, imaginary electronic SLDs, nuclear SLDs, magnetic SLDs, reflectivity scale factors \(r_{\mathrm{sc}}\), \(\log_{10}\) backgrounds and, for PNR, the mixing coefficient \(\alpha\). Units: \(d_i\) and \(\sigma_i\) are in \(\mathrm{\AA}\), all SLDs are in \(10^{-6}\,\mathrm{\AA}^{-2}\), while \(r_{\mathrm{sc}}\), the \(\log_{10}\) backgrounds and \(\alpha\) are dimensionless. The plot ranges correspond to the prior bounds.}
  \label{fig:PtFe_XRR_and_PNR_cornerplot_wider}
\end{figure}

\clearpage

\begin{table}[t]
\centering
\caption{Training ranges for the XRR model with standard slab parameterization}
\label{tab:training_ranges_XRR_standard}
\begin{tabular}{lll}
\toprule
\textbf{Parameter type} & \textbf{Training range} & \textbf{Physical unit} \\
\midrule
Thickness  & $[0,\;500]$     & \AA \\
Roughnesses  & $[0,\;60]$      & \AA \\
SLDs & $[0,\;50]$      &  10$^{-6}$ \AA$^{-2} $ \\ 
q-shift & $[-0.002,\;0.002]$ & \AA$^{-1}$ \\
Intensity scale       & $[0.9,\;1.1]$   & -  \\
\bottomrule
\end{tabular}
\end{table}

\begin{table}[t]
\centering
\caption{Training ranges for the NR model with standard slab parameterization}
\label{tab:training_ranges_NR_standard}
\begin{tabular}{llll}
\toprule
\textbf{Parameter type} & \textbf{Training range} & \textbf{Physical unit} \\
\midrule
Thicknesses  & $[0,\;1500]$     & \AA \\
Roughnesses  & $[0,\;60]$      & \AA \\
SLDs & $[-8,\;16]$      &  10$^{-6}$ \AA$^{-2} $ \\ 
Intensity scale       & $[0.9,\;1.1]$   & -  \\
Background (\(\log_{10}\))    & $[-10,\;-4]$ & - \\
\bottomrule
\end{tabular}
\end{table}

\begin{table}[t]
\centering
\caption{Training ranges for the NR model with parameterization incorporating solvent penetration}
\label{tab:training_ranges_NR_solvated}
\begin{tabular}{llll}
\toprule
\textbf{Parameter type} & \textbf{Training range} & \textbf{Physical unit} \\
\midrule
Thicknesses  & $[0,\;500]$     & \AA \\
Roughnesses  & $[0,\;60]$      & \AA \\
SLDs & $[-8,\;16]$      &  10$^{-6}$ \AA$^{-2} $ \\ 
Solvent volume fractions   & $[0,\;1]$     & -  \\
Intensity scale       & $[0.9,\;1.1]$   & -  \\
Background (\(\log_{10}\))    & $[-10,\;-4]$ & - \\
\bottomrule
\end{tabular}
\end{table}

\begin{table}[t]
\centering
\caption{Training ranges for the XRR model with absorption parameterization}
\label{tab:training_ranges_XRR_absorption}
\begin{tabular}{llll}
\toprule
\textbf{Parameter type} & \textbf{Training range} & \textbf{Physical unit} \\
\midrule
Thicknesses  & $[1,\;500]$     & \AA \\
Roughnesses  & $[0,\;20]$      & \AA \\
Re. SLDs & $[0,\;150]$      &  10$^{-6}$ \AA$^{-2} $ \\ 
Im. SLDs & $[0,\;30]$      &  10$^{-6}$ \AA$^{-2} $ \\ 
Intensity scale       & $[0.8,\;1.2]$   & -  \\
Background (\(\log_{10}\))    & $[-10,\;-4]$ & - \\
\bottomrule
\end{tabular}
\end{table}

\begin{table}[t]
\centering
\caption{Training ranges for the non--spin-flip PNR model}
\label{tab:training_ranges_PNR_nsf}
\begin{tabular}{llll}
\toprule
\textbf{Parameter type} & \textbf{Training range} & \textbf{Physical unit} \\
\midrule
Thicknesses  & $[1,\;500]$     & \AA \\
Roughnesses  & $[0,\;20]$      & \AA \\
Nuclear SLDs & $[0,\;10]$      &  10$^{-6}$ \AA$^{-2} $ \\ 
Magnetic SLDs & $[0,\;10]$      &  10$^{-6}$ \AA$^{-2} $ \\ 
Channel mixing factor & $[0,\;0.2]$      &  - \\ 
Intensity scale       & $[0.9,\;1.1]$   & -  \\
Background (\(\log_{10}\))    & $[-10,\;-4]$ & - \\
\bottomrule
\end{tabular}
\end{table}

\begin{table}[t]
\centering
\caption{Training ranges for the full spin-resolved PNR model}
\label{tab:training_ranges_PNR_sf}
\begin{tabular}{llll}
\toprule
\textbf{Parameter type} & \textbf{Training range} & \textbf{Physical unit} \\
\midrule
Thicknesses  & $[1,\;500]$     & \AA \\
Nuclear SLDs & $[0,\;10]$      &  10$^{-6}$ \AA$^{-2} $ \\ 
Magnetic SLDs & $[0,\;10]$      &  10$^{-6}$ \AA$^{-2} $ \\ 
Magnetization angles & $[0,\;180]$      &  degrees \\ 
\bottomrule
\end{tabular}
\end{table}

\begin{figure}[h]
  \centering
  \includegraphics[width=12cm]{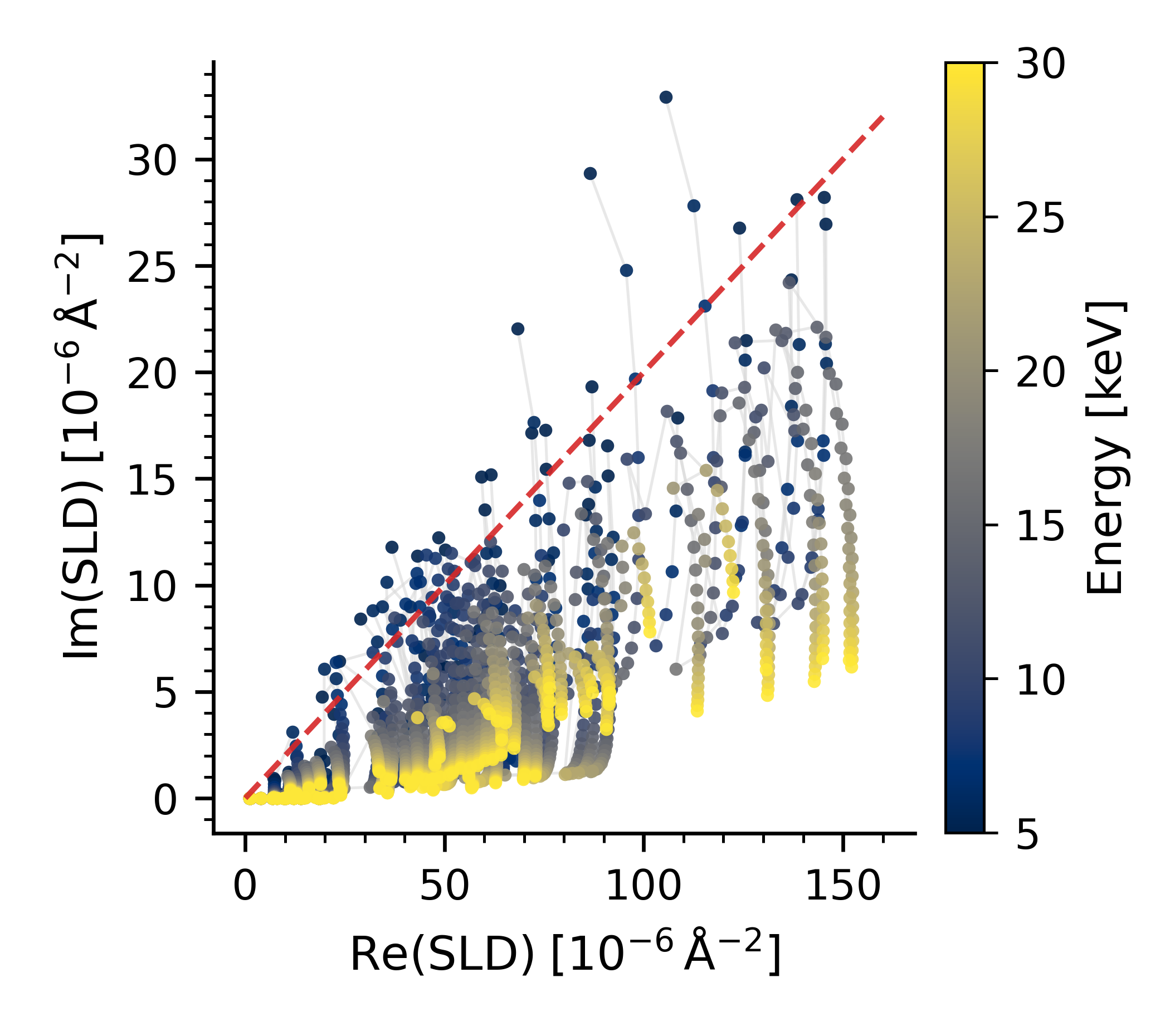}
  \caption{Relation between the real and imaginary SLD components for different elements in the periodic table and different X-ray beam energies computed using the \texttt{periodictable} package. The dashed line indicates the threshold used to restrict the training distribution of absorption parameters.}
  % \cite{PaulKienzle2024}
  \label{fig:absorption_sld}
\end{figure}

\end{document}